\documentclass{article}

 \usepackage[preprint]{neurips_2026}

\usepackage[utf8]{inputenc} % allow utf-8 input
\usepackage[T1]{fontenc}    % use 8-bit T1 fonts
\usepackage{hyperref}       % hyperlinks
\usepackage{url}            % simple URL typesetting
\usepackage{booktabs}       % professional-quality tables
\usepackage{amsfonts}       % blackboard math symbols
\usepackage{nicefrac}       % compact symbols for 1/2, etc.
\usepackage{microtype} 
\usepackage{marvosym}% microtypography
\usepackage{xcolor}         % colors
\usepackage{amsmath}
\usepackage{amssymb}
\usepackage{bm}
\usepackage{graphicx}
\usepackage{array}
\usepackage{colortbl}
\usepackage{tcolorbox}
\usepackage{multirow}
\usepackage{float}
\usepackage{enumitem}
\usepackage{pifont}

\tcbuselibrary{breakable, skins}  % optional but useful
\newcommand{\methodname}{\textsc{Arro}}

\title{Scalable Minimal-Change Learning for \\ Controllable Image Editing}

\author{%
  Shuo Chen$^{1}$~~, 
  Fengming Huang$^{1}$, 
  Yu Yao$^{1}$\textsuperscript{\Letter}, 
  Mingming Gong$^{2}$, 
  Tongliang Liu$^{1}$ \\
  $^{1}$Sydney AI Centre, The University of Sydney \\
  $^{2}$University of Melbourne \\
  \texttt{yu.yao@sydney.edu.au} \\
}

\begin{document}

\maketitle

\begin{abstract}
Image editing aims to modify specific attributes of an image while preserving all other aspects. In practice, however, applying even simple editing instructions to existing methods often leads to unintended changes. 
We identify the central issue: a fundamental causal principle of \emph{minimal change}, which requires that an intervention alter only the intended attributes in the output while leaving all others invariant, has not been explicitly incorporated as an optimization objective. 
To encourage minimal change, prior work rooted in causal representation learning typically imposes an $L_1$ regularizer on the latent difference between pre- and post-edit representations.
However, this strategy does not scale to modern image editing models: $L_1$ regularization fails to induce true sparsity and instead tends to shrink all latent differences uniformly; sparsity in the latent space does not translate to localized change in the output of nonlinear deep image generation models; and the multi-domain or counterfactual supervision it requires is generally unavailable.
To effectively leverage the minimal change principle for controllable image editing, we cast it as a reinforcement learning objective. Rather than constraining latent representations, we design rewards that directly capture minimal change over intervention outcomes and optimize them explicitly during training.  
To produce these rewards reliably, we further introduce an agentic reward model that leverages the chain-of-thought reasoning ability of vision-language models without requiring human annotations. The idea is that instead of emitting a single, unreliable scalar score, the model reasons explicitly about two complementary failure modes, unimplemented changes and unintended changes, and produces this supervision automatically.
Across four benchmarks, ARRO improves average EditScore from 5.21 to 5.88. A region-aware evaluation on 600 examples further shows an 8.4\% reduction in off-target pixel change relative to the base editor. Controlled reward and SFT comparisons, blinded human evaluations, and transfer to OmniGen2 provide complementary evidence for the reward design and its applicability beyond one backbone.
\textbf{Code:} \url{https://github.com/Showwwwwwwww/ARRO}.
\end{abstract}

\section{Introduction}
\label{sec:intro}

\begin{figure}[t]
    \centering
    \setlength{\abovecaptionskip}{4pt}
    \includegraphics[width=\textwidth]{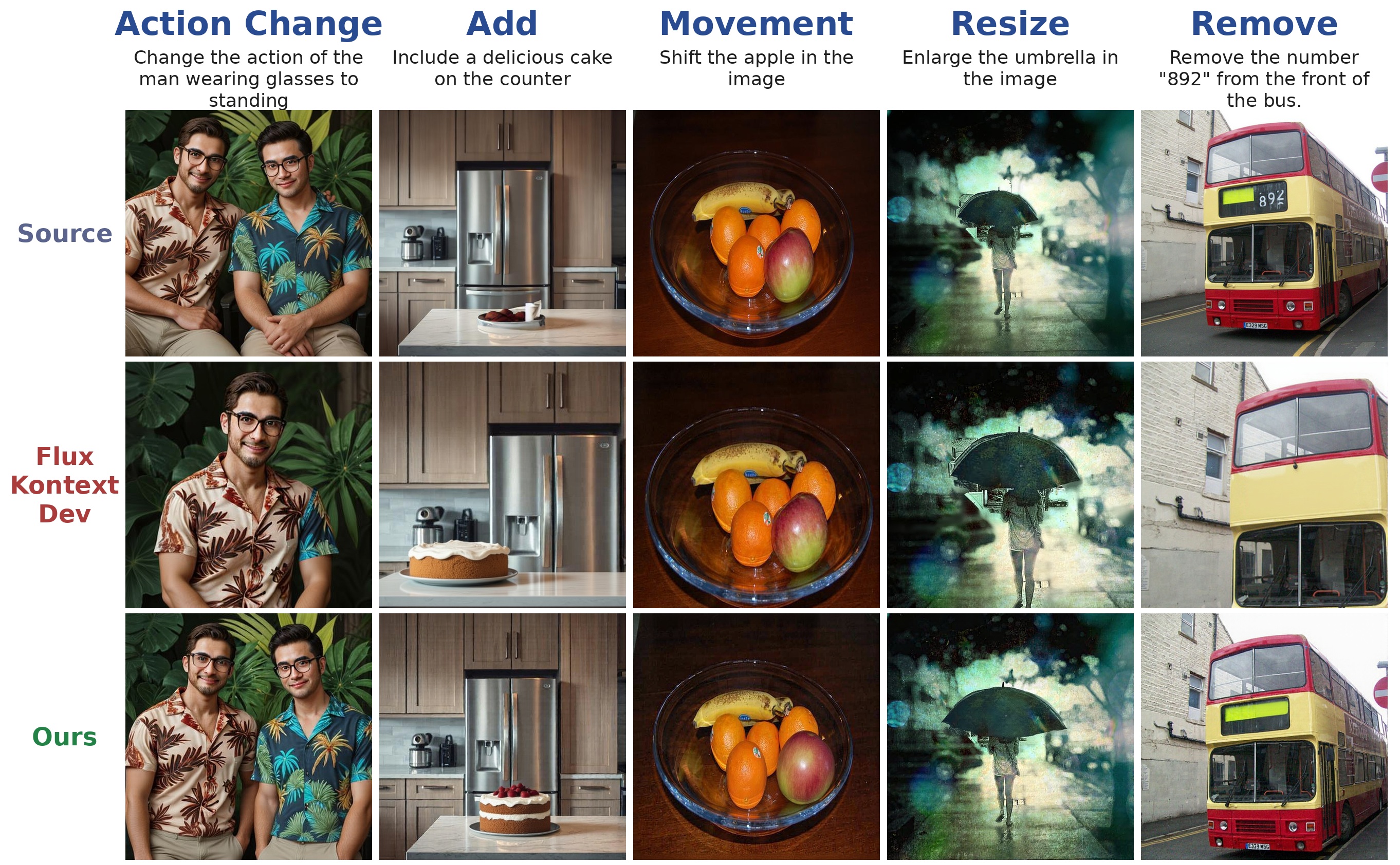}
    \caption{
    Illustrative edits by FLUX.1~Kontext-dev and \methodname{}. ARRO better preserves the companion, fruit arrangement, umbrella scene, and bus surroundings while carrying out the requested edits. The ambiguous ``Include a delicious cake'' example remains imperfect: ARRO adds a cake but replaces the original dish. It is retained as a visible failure case rather than evidence of perfect preservation.}
    \label{fig:teaser}
\end{figure}
\vspace{-1.0em}

% ===== ¶1: the problem =====
Image editing aims to modify specific attributes of an image while preserving all other aspects. Recent instruction-based editors~\citep{instructpix2pix,magicbrush,emuedit,flux_kontext,qwen_image_edit,omnigen2,bagel,icedit} have made impressive progress on visual quality and instruction following. Yet even simple instructions still trigger widespread, unintended changes: ``add glasses'' subtly alters the subject's hair and skin; ``change the car's color'' redraws the background; ``remove the sign'' shifts the lighting of the entire scene. This is not a cosmetic issue but a fundamental one: it determines whether a generative model can be trusted as a controllable editing tool, rather than a high-fidelity sampler that happens to follow instructions most of the time.

% ===== ¶2: minimal change -> three obstacles =====
We identify a central yet under-explored issue behind this gap: the fundamental causal principle of \emph{minimal change}, which requires that an intervention alter only the intended attributes in the output while leaving all others invariant, has not been explicitly incorporated as an optimization objective in modern image editing models. 

Minimal change is widely applied in causal representation learning~\citep{scholkopf_toward_2021,peters_elements_2017,perry_smshift_2022,lopez_smshift_2022}. Existing methods (e.g., \citep{causalvae,dear,scmvae}) operationalize it by imposing an
$L_1$ regularizer on the latent difference between pre- and post-intervention representations, encouraging interventions on a single latent factor to leave the remaining factors unchanged. This strategy is difficult to adapt effectively to modern image editing models, due to three obstacles.
\textbf{(O1)} $L_1$ does not reliably induce sparsity: it tends to shrink all latent coordinates uniformly toward small values rather than driving most of them to exactly zero~\citep{ppe_sparse_2025}, so latent edits remain dense, just quieter. \textbf{(O2)} Latent sparsity does not translate to output locality: modern decoders are large, deeply nonlinear neural networks with a latent-to-image Jacobian that is far from axis-aligned, so even a sparse change in the latent space can propagate into broad, unintended modifications across the output image~\citep{interfacegan,stylegan2}. \textbf{(O3)} The required supervision does not scale: causal representation learning methods rely on multi-domain datasets with attribute labels or paired counterfactual examples, which are expensive or generally unavailable.

% ===== ¶3: method overview =====
We address all three obstacles by reformulating minimal change as a reinforcement learning (RL) objective on intervention outcomes rather than latents. Designing rewards that directly capture minimal change on the output dissolves \textbf{(O1)} and \textbf{(O2)} simultaneously: there is no $L_1$ approximation to optimize, and the decoder's nonlinearity is no longer a problem, because the reward is computed directly on the output rather than on the latent that produces it. Concretely, the supervised dataset contains only source-instruction pairs $(\bm{x}_{\text{src}},\, c)$; during RL, the current policy samples candidate edits $\bm{x}_i$, which are then assigned minimal-change rewards $r_i$ according to how well they implement the instruction while leaving everything else intact.

% ===== ¶4: agentic reward (addresses O3) =====
To address \textbf{(O3)}, a minimal-change reward has to be reliably designed without human annotation. A natural choice would be to ask a vision-language model (VLM) for a single editing quality score, as in recent reward-modeling work~\citep{onereward, editscore}. However, directly prompting a VLM for a numerical score is unreliable. First, the score is poorly calibrated: scalar VLM ratings are noisy and high-variance, with thresholds that shift across images, so the same numerical score can mean different things on different inputs~\citep{thinkrl_edit_2026}. Second, the score is uninformative about the type of failure: a minimal-change reward must capture two distinct aspects, whether the instructed edit was carried out and whether everything else was left alone, but a single number collapses both into one signal, so the policy cannot tell which aspect it is failing on.

To make the minimal-change reward more reliable, we design an agentic reward model: rather than asking the VLM for a numerical score, we leverage its chain-of-thought reasoning ability to perform a structured audit of the edit. We decompose the minimal-change reward into two complementary components: \emph{unimplemented changes}, where the instructed edit is missing or incomplete, and \emph{unintended changes}, where modifications appear outside the instructed scope. For each component, the VLM reasons over the editing triplet of source image, instruction, and edited output, and produces a list of identified issues. To stabilize the agentic reward model, we further introduce a \emph{rubric refinement} loop that enforces consistency within each GDPO group (the set of candidate edits sampled for a given source-instruction pair). An LLM first merges the candidates' audit lists into a single deduplicated rubric. The VLM then performs a second-pass evaluation in which, for each candidate, it selects the applicable items from this shared rubric, and the two reward signals are computed from the selected items.

% ===== ¶5: results =====
On top of FLUX.1~Kontext, our method lifts average EditScore from $5.21$ to $5.88$ across four benchmarks (MinEval, MagicBrush, AnyBench, and Emu-Edit). A masked pixel-difference evaluation on 600 examples measures 8.4\% less off-target change than the base editor. Under fixed GDPO training, replacing ARRO's rewards with EditScore Prompt-Following and Consistency rewards yields $3.78$ average EditScore, while ARRO achieves $5.88$. On MinEval, ARRO-GDPO also exceeds reward-guided Top-10 SFT ($5.669$ vs.\ $4.629$ EditScore). Blinded human comparisons and transfer to OmniGen2 provide further evidence. These results support supervising the \emph{effect} of an intervention.

% ===== contributions =====
\paragraph{Contributions.}
\textbf{(1)}~We diagnose three reasons why latent-space sparsity fails to deliver minimal change in modern generative editors ineffective $L_1$ regularization, decoder nonlinearity, and the lack of scalable supervision and recast minimal change as a reinforcement learning objective on intervention outcomes, optimized end-to-end over a flow-matching editor.
\textbf{(2)}~We design an agentic reward model that, unlike scalar VLM judges, reasons over the editing triplet to produce decomposed efficacy and locality rewards from raw images alone counting unimplemented and unintended changes, respectively stabilized by a group-level rubric refinement step that anchors all candidates to a shared, deduplicated set of canonical issues.
\textbf{(3)}~We developed MinEval, a dataset of automatically generated edit prompts and source images covering atomic changes across six edit types, with disjoint training and held-out evaluation splits. MinEval is designed to facilitate research into controllable image editing.
\textbf{(4)}~On top of FLUX.1~Kontext, \methodname{} improves average EditScore from $5.21$ to $5.88$ across four benchmarks and reduces masked off-target pixel change by 8.4\% on a 600-example evaluation. Reward-pipeline, EditScore-reward, and Top-10 SFT controls, human evaluations, and OmniGen2 transfer probe the source and breadth of these gains (Sec.~\ref{sec:exp}; App.~\ref{app:rebuttal_results}).

\section{Related Work}
\label{sec:related}

\paragraph{Instruction-based Image Editing.} 
Following \citet{instructpix2pix}, recent models like FLUX.1 Kontext \citep{flux_kontext} treat locality as an emergent property of pretraining. However, unintended background drift persists in current benchmarks \citep{gie_bench, gedit_bench_v2}. Post-training approaches address this via preference optimization \citep{diffusion_dpo, rpo} or online RL \citep{flowgrpo, thinkrl_edit_2026} using learned reward models \citep{editscore, onereward}. We depart from this line in two ways: (1) we argue human preference rewards often conflate aesthetics with adherence rather than penalizing non-minimal changes, and (2) our agentic VLM reward is zero-shot, bypassing the need for expensive, learned reward-model training.

\paragraph{Flow Matching and Sampling Efficiency.}
Research on generative flow matching has improved few-step synthesis through model-aligned training couplings~\citep{lin2025modelalignedcoupling} and sub-mode conditioning for diverse one-step generation~\citep{lin2026subflow}. In a distinct, discrete-diffusion setting, parallel-in-time sampling accelerates the sequential generation process~\citep{yao2026parallelintime}. These works address transport, diversity, and sampling cost; our objective concerns the scope of changes made by an instruction-conditioned editor.

\paragraph{The Minimal Change Principle.} 
The principle that interventions should perturb only targeted mechanisms originates in causal inference \citep{pearl_causality, scholkopf_toward_2021}. In generative modeling, this is often operationalized through structured latent spaces \citep{causalvae, causaldiffae} or training-free latent constraints \citep{saedit_2025, ppe_sparse_2025}. While we share the conceptual goal of sparse interventions, we argue that enforcing minimal change on a latent passed through a non-linear decoder is a brittle proxy. Instead, we optimize for minimal change directly in the output space via visual auditing.

\paragraph{Task-specific representations and metrics.}
EditCLIP~\citep{editclip} learns representations of source-to-edit transformations for exemplar-based editing and evaluation; NearID~\citep{nearid} uses matched-background distractors to learn identity representations; StyleID~\citep{styleid} learns a facial-identity metric from perceptual judgments across stylizations. These methods use supervision tailored to a predefined representation or metric. ARRO instead constructs instruction-specific, decomposed reward signals from candidate edits without per-instruction human labels; the resulting issue annotations can also support offline candidate selection. We do not claim that these representation-learning tasks are interchangeable with online editing rewards.

\paragraph{Safety and Asset Protection.}
Adjacent work examines visual-instruction jailbreaks of image-to-video models~\citep{zheng2026vii}, conflict-aware safety guidance for text-to-image diffusion~\citep{xiang2026safetycollides}, protection of {3D} Gaussian splatting assets from instruction-driven editing~\citep{hong2025adlift}, and defenses against image-to-video generation and fine-tuning-based video customization~\citep{huang2026temporalprotection}. These studies address misuse and unauthorized modification; our focus is preserving untargeted content when carrying out an authorized image edit.

\paragraph{Agentic VLM Rewards.} 
Standard VLM-based rewards typically provide scalar scores \citep{vlm_rm, onereward}. Recent work has introduced structured reasoning or binary checklists to improve interpretability \citep{thinkrl_edit_2026, rewarddance}. Our approach sits at the intersection of agentic VLMs \citep{vtool_r1} and RL; rather than a global similarity score, our reward is derived from a structured audit of ``unimplemented'' versus ``unintended'' changes. This provides a granular, per-region optimization signal that makes locality a directly optimizable property. 

Due to the limited space, the extended related work is provided in Appendix~\ref{app:ref}.

\section{Method}
\label{sec:method}

We cast \emph{minimal change} as a reinforcement learning objective in the output space of a generative editor: given a source image $\bm{x}_{\text{src}}$ and an editing instruction $c$, the policy $\pi_\theta$ should realize the instructed effect while leaving every other aspect of the output unchanged. This re-frames controllable image editing away from representation-level sparsity, which constrains \emph{how} the model encodes factors, and toward outcome-level optimization, which constrains \emph{what the output looks like} after the intervention. An outcome-level reward must therefore observe two complementary failure modes at once: \emph{unimplemented changes}, where the instructed edit is missing or incomplete, and \emph{unintended changes}, where modifications appear outside the instructed scope.

\begin{figure}[t]
    \centering
    \includegraphics[width=\textwidth]{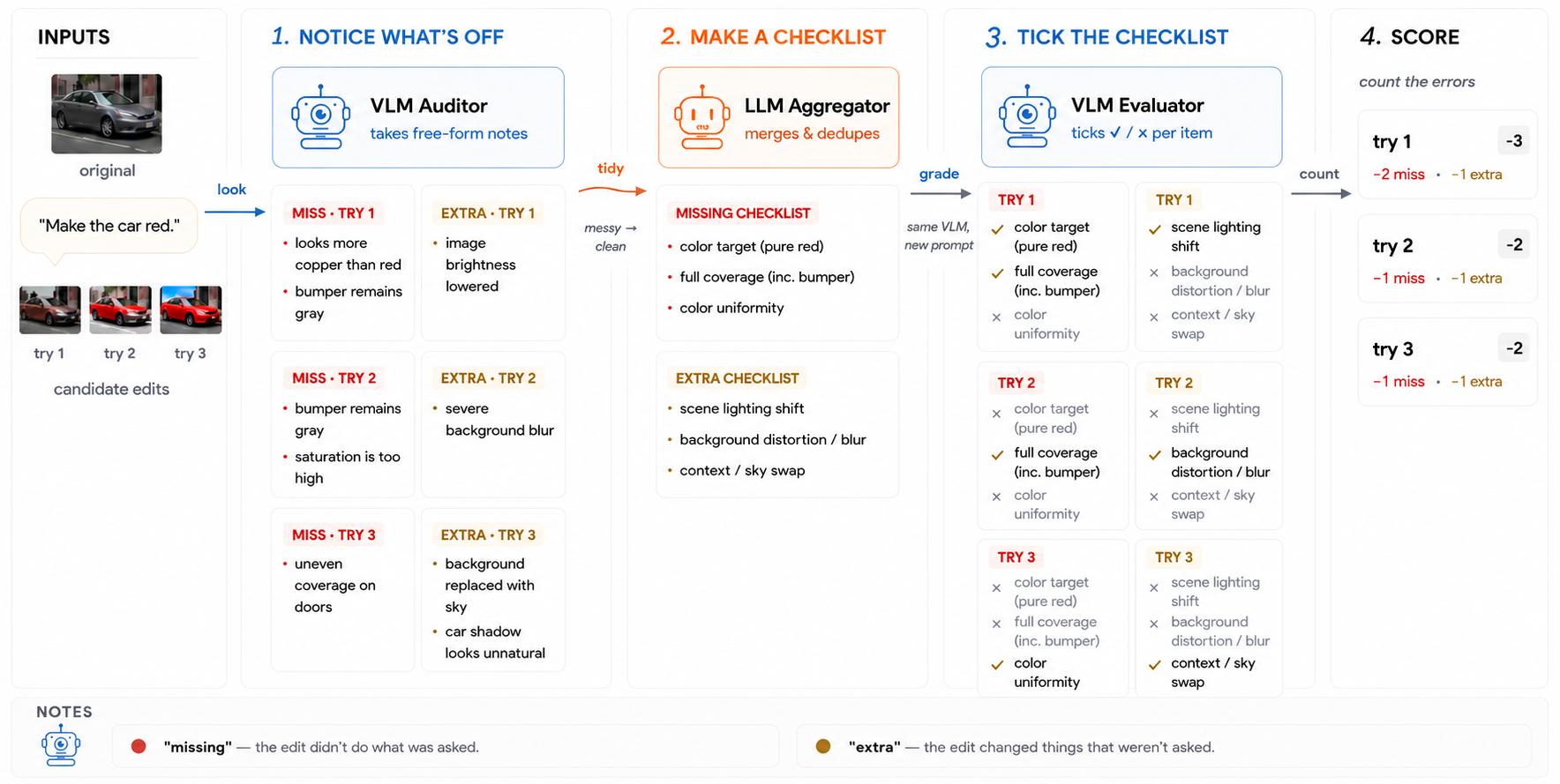}
    \caption{Overview of our agentic reward pipeline. For each editing triplet $(\bm{x}_{\text{src}}, c, \bm{x}_i)$, the vision-language model $\Phi_{\text{V}}$ performs a chain-of-thought audit, producing two raw lists of unimplemented and unintended changes. An LLM $\Phi_{\text{L}}$ aggregates these lists across the GDPO group into deduplicated rubrics, and $\Phi_{\text{V}}$ re-evaluates each candidate against the shared rubrics. The resulting two scalar rewards $(r^{(1)}_i, r^{(2)}_i)$ are each group-normalized into per-channel advantages $(A^{(1)}_i, A^{(2)}_i)$, which are then combined into the final advantage that drives the flow-based GDPO update.}
    \label{fig:pipeline}
\end{figure}

Figure~\ref{fig:pipeline} illustrates the pipeline. Our method \methodname{} realizes this view in two stages: an agentic VLM audits each candidate edit and decomposes it into unimplemented and unintended lists (Sec.~\ref{sec:dualtrack}); a rubric refinement step then makes these audits group-consistent and yields two scalar rewards $r^{(1)}_i, r^{(2)}_i$ commensurable across the group (Sec.~\ref{sec:rubric}). The two channels are each group-normalized, their advantages combined, and the result drives a flow-based GDPO update (Sec.~\ref{sec:opt}) on top of FLUX.1~Kontext~\citep{flux_kontext}. 

\paragraph{Notation.} Source $\bm{x}_{\text{src}}\!\in\!\mathcal{X}$, instruction $c$, dataset $\mathcal{D}\!=\!\{(\bm{x}_{\text{src}}^{(k)}, c^{(k)})\}$, flow-based policy $\pi_\theta(\bm{x}\!\mid\!\bm{x}_{\text{src}},c)$. The reward model invokes a VLM $\Phi_{\text{V}}$ and a text-only LLM $\Phi_{\text{L}}$ through three fixed prompts: audit $\mathsf{P}_{\text{audit}}$, aggregation $\mathsf{P}_{\text{agg}}$, verification $\mathsf{P}_{\text{verify}}$ (full templates in App.~\ref{app:prompts}).

\subsection{Preliminaries}
\label{sec:prelim}

\paragraph{Rectified-flow editing model.}
FLUX.1~Kontext is a rectified-flow transformer~\citep{rectified_flow,flux_kontext} that conditions on a reference image by concatenating its VAE latent $\bm{z}_{\text{src}}$ with the noisy latent $\bm{z}_t$ along the sequence axis. With $\bm{z}_0\!\sim\!p_{\text{data}}$ (the target-image latent), $\bm{\epsilon}\!\sim\!\mathcal{N}(\bm{0},\bm{I})$, and $t\!\sim\!\mathcal{U}[0,1]$, the velocity field $\bm{v}_\theta$ is trained against the rectified-flow objective
\begin{equation}
\mathcal{L}_{\text{RF}}(\theta) \;=\; \mathbb{E}_{t,\,\bm{z}_0,\,\bm{\epsilon}}\!\left[\,\big\|\bm{v}_\theta\!\big([\bm{z}_t;\bm{z}_{\text{src}}],\,t,\,c\big) - (\bm{\epsilon}-\bm{z}_0)\big\|_2^2\,\right],\qquad \bm{z}_t=(1-t)\bm{z}_0+t\bm{\epsilon}.
\label{eq:rf}
\end{equation}
We perform RL on top of a pretrained editor; the flow-matching loss is not used during fine-tuning.

\paragraph{GDPO.}
GDPO~\citep{gdpo} adapts group-relative preference optimization to the multi-reward setting. For each input pair $(\bm{x}_{\text{src}},c)\!\sim\!\mathcal{D}$, it samples a group $\mathcal{G}=\{\bm{x}_1,\dots,\bm{x}_N\}$ from the current policy $\pi_{\theta_{\text{old}}}$ and assigns each candidate $n$ scalar rewards $\{r^{(k)}_i\}_{k=1}^{n}$, with one per reward channel. Rather than collapsing the channels into a single scalar before normalization, GDPO $z$-scores each channel separately within $\mathcal{G}$,
\begin{equation}
A^{(k)}_i \;=\; \frac{r^{(k)}_i - \mu^{(k)}_\mathcal{G}}{\sigma^{(k)}_\mathcal{G}+\varepsilon_0},\qquad k=1,\dots,n,
\label{eq:gdpo_adv_main}
\end{equation}
forms a (weighted) sum $A^{\text{sum}}_i \!=\! \sum_k w_k A^{(k)}_i$, and re-normalizes $A^{\text{sum}}_i$ across the batch to keep the advantage scale stable as $n$ grows. The policy is then updated by a per-step PPO-clip objective along an SDE discretization of the rectified flow~\citep{mixgrpo}. Full equations and the MixGRPO sampler are deferred to App.~\ref{app:gdpo_full}. 

\subsection{Agentic Reward Model via Chain-of-Thought Auditing}
\label{sec:dualtrack}

We deploy the vision-language model $\Phi_{\text{V}}$ as an autonomous \emph{causal observer}. Rather than asking the VLM $\Phi_{\text{V}}$ to regress a scalar score, which is unreliable for the calibration, variance, and conflation reasons discussed in Sec.~\ref{sec:intro}, we prompt it to perform a multi-step audit of the editing triplet $(\bm{x}_{\text{src}}, c, \bm{x}_i)$ for each candidate $\bm{x}_i\!\in\!\mathcal{G}$.

\paragraph{Per-candidate audit.}
For each candidate $\bm{x}_i$, the audit pass invokes the VLM $\Phi_{\text{V}}$ with the audit prompt $\mathsf{P}_{\text{audit}}$ instantiated on the editing triplet:
\begin{equation}
(\widetilde{\mathcal{U}}_i,\,\widetilde{\mathcal{O}}_i) \;=\; \Phi_{\text{V}}\!\big(\mathsf{P}_{\text{audit}}(\bm{x}_{\text{src}},\, c,\, \bm{x}_i)\big).
\label{eq:audit}
\end{equation}

\begin{tcolorbox}[
    title={Audit prompt $\mathsf{P}_{\text{audit}}$ (a summary)},
    fonttitle=\small\bfseries, fontupper=\small
]
\textbf{Inputs:} source image $\bm{x}_{\text{src}}$, edited image $\bm{x}_i$, instruction $c$. \\[2pt]
\textbf{Step 1.} Parse the instruction $c$: identify what should change and what should remain. \\[2pt]
\textbf{Step 2.} Compare $\bm{x}_{\text{src}}$ and $\bm{x}_i$ side by side; note all visible differences. \\[2pt]
\textbf{Step 3.} Emit two open-ended natural-language lists:
\begin{itemize}
    \item the raw unimplemented list $\widetilde{\mathcal{U}}_i$, enumerating aspects of $c$ not realized in $\bm{x}_i$;
    \item the raw unintended list $\widetilde{\mathcal{O}}_i$, enumerating visible changes outside the scope of $c$.
\end{itemize}
\textbf{Output:} JSON object containing $\widetilde{\mathcal{U}}_i$ and $\widetilde{\mathcal{O}}_i$.
\end{tcolorbox}

The two outputs are ordered lists $\widetilde{\mathcal{U}}_i = (u_{i,1},\,u_{i,2},\,\dots)$ and $\widetilde{\mathcal{O}}_i = (e_{i,1},\,e_{i,2},\,\dots)$, where each entry $u_{i,j}$ or $e_{i,j}$ is a short natural-language description. Crucially, $\Phi_{\text{V}}$ never outputs a number on a continuous scale; it only identifies and describes changes, both tasks modern VLMs handle reliably. However, raw counts $|\widetilde{\mathcal{U}}_i|$ and $|\widetilde{\mathcal{O}}_i|$ are unreliable across candidates; We address this in Sec.~\ref{sec:rubric}.   

\subsection{Rubric-Refined Reward}
\label{sec:rubric}

Counting raw audit lists directly is unreliable: free-form VLM enumerations duplicate the same artifact under different wordings, fragment a single change into several entries, and drift in granularity across candidates, so $|\widetilde{\mathcal{U}}_i|$ and $|\widetilde{\mathcal{O}}_i|$ are not commensurable across $\mathcal{G}$. To fix this, we introduce the language model $\Phi_{\text{L}}$ as a \emph{rubric aggregator} operating within each GDPO group.

\paragraph{Aggregation.}
The aggregator is invoked once per group. Its prompt $\mathsf{P}_{\text{agg}}$ receives both raw issue collections and returns one rubric for each failure mode:
\begin{equation}
\big(\mathcal{R}^{\mathcal{U}}_\mathcal{G},\,\mathcal{R}^{\mathcal{O}}_\mathcal{G}\big)
\;=\; \Phi_{\text{L}}\!\Big(\mathsf{P}_{\text{agg}}\big(c,\,\operatorname{concat}_{i=1}^{N}\widetilde{\mathcal{U}}_i,\,\operatorname{concat}_{i=1}^{N}\widetilde{\mathcal{O}}_i\big)\Big).
\label{eq:rubric}
\end{equation}

\begin{tcolorbox}[
    title={Aggregation prompt $\mathsf{P}_{\text{agg}}$ (a summary)},
    fonttitle=\small\bfseries, fontupper=\small
]
\textbf{Inputs:} instruction $c$ and both collections of raw audit lists across the group: $\operatorname{concat}_{i=1}^{N}\widetilde{\mathcal{U}}_i$ and $\operatorname{concat}_{i=1}^{N}\widetilde{\mathcal{O}}_i$. \\[2pt]
\textbf{Step 1 \emph{(instruction filter).}} Filter changes requested or directly implied by $c$ out of the unintended collection; retain the unimplemented collection. \\[2pt]
\vspace{2pt}   
\textbf{Step 2.} Consolidate the entries: merge entries describing the same underlying issue under different wordings, split entries that conflate distinct issues, normalize the granularity of all entries to a single atomic visually-checkable change, and drop vague, speculative, or non-visual entries. \\[2pt]
\textbf{Step 3.} Produce two separate deduplicated rubrics of canonical issues, written in neutral form so that any candidate can be checked against each entry. \\[2pt]
\textbf{Output:} both $\mathcal{R}^{\mathcal{U}}_\mathcal{G}$ and $\mathcal{R}^{\mathcal{O}}_\mathcal{G}$ in one response.
\end{tcolorbox}

The output unimplemented rubric $\mathcal{R}^{\mathcal{U}}_\mathcal{G} = \{u^*_1,\dots,u^*_{M_\mathcal{U}}\}$ and unintended rubric $\mathcal{R}^{\mathcal{O}}_\mathcal{G} = \{e^*_1,\dots,e^*_{M_\mathcal{O}}\}$ are deduplicated lists whose entries we call \emph{canonical issues}. Together, the two rubrics form a per-group, group-consistent vocabulary against which every candidate in $\mathcal{G}$ can be evaluated on equal footing.

\paragraph{Rubric-anchored verification.}
A second VLM pass re-evaluates each candidate against both group rubrics in one verification call:
\begin{equation}
\big(\mathcal{U}_i,\,\mathcal{O}_i\big) \;=\; \Phi_{\text{V}}\!\big(\mathsf{P}_{\text{verify}}(\bm{x}_{\text{src}},\,c,\,\bm{x}_i,\,\mathcal{R}^{\mathcal{U}}_\mathcal{G},\,\mathcal{R}^{\mathcal{O}}_\mathcal{G})\big).
\label{eq:verify}
\end{equation}

\begin{tcolorbox}[
    title={Verification prompt $\mathsf{P}_{\text{verify}}$ (a summary)},
    fonttitle=\small\bfseries, fontupper=\small
]
\textbf{Inputs:} source image $\bm{x}_{\text{src}}$, edited image $\bm{x}_i$, instruction $c$, and both rubrics $\mathcal{R}^{\mathcal{U}}_\mathcal{G}$ and $\mathcal{R}^{\mathcal{O}}_\mathcal{G}$. \\[2pt]
\textbf{Step 1.} Parse the instruction $c$ (identical to Step 1 of $\mathsf{P}_{\text{audit}}$). \\[2pt]
\textbf{Step 2.} Compare $\bm{x}_{\text{src}}$ and $\bm{x}_i$ side by side (identical to Step 2 of $\mathsf{P}_{\text{audit}}$). \\[2pt]
\textbf{Step 3.} For each canonical issue in each rubric, return a Boolean verdict on whether that issue is visibly present in $\bm{x}_i$. No new issues are admitted; the wording of rubric items is not modified. \\[2pt]
\textbf{Output:} both verified subsets $\mathcal{U}_i\!\subseteq\!\mathcal{R}^{\mathcal{U}}_\mathcal{G}$ and $\mathcal{O}_i\!\subseteq\!\mathcal{R}^{\mathcal{O}}_\mathcal{G}$ in one response.
\end{tcolorbox}

The verification prompt $\mathsf{P}_{\text{verify}}$ shares its first two reasoning steps with the audit prompt $\mathsf{P}_{\text{audit}}$, that is, parsing the instruction $c$ and comparing $\bm{x}_{\text{src}}$ with $\bm{x}_i$, but its third step replaces open-ended generation with closed-form selection over the input rubric. Anchoring all candidates in the group $\mathcal{G}$ to the same canonical sets removes the duplication, fragmentation, and granularity drift that contaminated the raw audit outputs in~\eqref{eq:audit}.

\paragraph{Reward derivation.}
The two reward channels are simple counts over the verified rubric items:
\begin{equation}
r^{(1)}_i \;=\; -\,|\mathcal{U}_i|,
\qquad
r^{(2)}_i \;=\; -\,|\mathcal{O}_i|.
\label{eq:rewards}
\end{equation}
Both rewards are negated so that fewer issues correspond to higher reward, and both are integer-valued, with the unimplemented reward $r^{(1)}_i\!\in\!\{-M_\mathcal{U},\dots,0\}$ bounded by the size of the unimplemented rubric $\mathcal{R}^{\mathcal{U}}_\mathcal{G}$, and similarly for the unintended reward $r^{(2)}_i\!\in\!\{-M_\mathcal{O},\dots,0\}$. Because every candidate in the group $\mathcal{G}$ is scored against the same canonical sets $\mathcal{R}^{\mathcal{U}}_\mathcal{G}$ and $\mathcal{R}^{\mathcal{O}}_\mathcal{G}$, the counts in~\eqref{eq:rewards} are directly comparable across the group, which is the property the advantage estimator in Sec.~\ref{sec:opt} relies on.

\subsection{Learning Objective}
\label{sec:opt}

Following GDPO, we instantiate the multi-reward pipeline of Eq.~\eqref{eq:gdpo_adv_main} with $n\!=\!2$ channels. We $z$-score $r^{(1)}_i$ and $r^{(2)}_i$ within the group $\mathcal{G}$ to obtain per-channel advantages $A^{(1)}_i$ and $A^{(2)}_i$, combine them linearly into a group-level advantage
\begin{equation}
A^{\text{sum}}_i \;=\; w_1\,A^{(1)}_i \;+\; w_2\,A^{(2)}_i,\qquad w_1, w_2 > 0,
\label{eq:our_advantage_sum}
\end{equation}
and re-normalize across the training batch $\mathcal{B}$ (all candidates from all groups in the batch),
\begin{equation}
\hat{A}_i \;=\; \frac{A^{\text{sum}}_i \;-\; \mu^{\text{sum}}_\mathcal{B}}{\sigma^{\text{sum}}_\mathcal{B}+\varepsilon_0},
\label{eq:our_advantage}
\end{equation}
with $\mu^{\text{sum}}_\mathcal{B},\sigma^{\text{sum}}_\mathcal{B}$ the mean and standard deviation of $A^{\text{sum}}$ across $\mathcal{B}$. Per-channel group normalization places the unimplemented and unintended signals on a common, dimensionless scale within each group, so neither dominates the gradient simply by virtue of having larger raw variance, and the relative emphasis between efficacy and locality is controlled by $(w_1, w_2)$ rather than implicitly by reward-scale differences. The batch-level renormalization in Eq.~\eqref{eq:our_advantage} keeps the numerical scale of $\hat{A}_i$ stable across batches and consistent with the single-reward baseline, so the PPO clipping threshold $\epsilon_{\text{clip}}$ behaves uniformly. Collapsing the two rewards into a single scalar \emph{before} group normalization would forfeit the per-channel scaling property; we show in Sec.~\ref{sec:exp_ablation} (variant 2) that doing so degrades training by ${\sim}1.5$ EditScore. Plugging $\hat{A}_i$ into the GDPO PPO-clip objective gives our final training objective $\mathcal{J}_{\text{ours}}(\theta)$ (full form in App.~\ref{app:gdpo_full}). Groups with zero within-group variance on both channels supply no learnable ranking and are skipped at the gradient step, but are largely eliminated upstream by the rejection-sampling stage of our data pipeline (App.~\ref{app:data_reject}).

\section{Experiments}
\label{sec:exp}

\subsection{Setup}
\label{sec:exp_setup}

\textbf{Training data.} We use the $\sim$10K-prompt training split of \textbf{MinEval}, a dataset we construct by a three-stage VLM pipeline over MagicBrush~\citep{magicbrush} source images, covering six atomic edit types (\texttt{add}, \texttt{remove}, \texttt{move}, \texttt{replace}, \texttt{appearance\_change}, \texttt{action\_change}) and rejection-sampled with the un-tuned base editor to drop unrealizable and trivial prompts; full pipeline, prompts, and checklist are in App.~\ref{app:data_construction}.

\textbf{Evaluation benchmarks.} We evaluate on four \emph{complementary} benchmarks: \textbf{(1)}~the held-out 600-prompt split of \textbf{MinEval} (in-distribution; 6 atomic edit types, disjoint from training), \textbf{(2)}~\textbf{MagicBrush}~\citep{magicbrush} for human-curated edits from a different pipeline (535 prompts), \textbf{(3)}~\textbf{AnyBench}~\citep{anyedit} for breadth across 18 edit types with fine-grained per-type analysis (900 prompts), and \textbf{(4)}~\textbf{Emu-Edit}~\citep{emuedit} for out-of-distribution robustness at scale (3{,}589 prompts; 8 task types). Per-type analysis is concentrated on AnyBench, which is the only benchmark with edit-type labels at this granularity.

\textbf{Metrics.} The primary metric is \textbf{EditScore-Overall (ES)}~\citep{editscore}, a Qwen3-VL-8B$+$LoRA edit-quality judge on $[0,10]$ ($\uparrow$). We complement it with two off-the-shelf preference scorers from different model families: \textbf{HPSv2 (HPS)}~\citep{hpsv2}, a CLIP-H based score on $[0,1]$ ($\uparrow$), and \textbf{ImageReward (IR)}~\citep{imagereward}, a BLIP-based real-valued score ($\uparrow$). To measure preservation more directly, we also report \textbf{OffTarget-L1} ($\downarrow$), the mean absolute source-to-edit pixel difference outside a SAM~2 target mask on 600 randomly selected examples from the four benchmarks; its definition and scope are in App.~\ref{app:offtarget}. The three EditScore sub-scores (PF, Cons, PQ) are in App.~\ref{app:metric_subscores}.

\textbf{Baselines.} We compare against six external instruction-edit models InstructPix2Pix~\citep{instructpix2pix}, MagicBrush~\citep{magicbrush}, OmniGen2~\citep{omnigen2}, UniWorld-v1~\citep{lin2025uniworldhighresolutionsemanticencoders}, ICEdit~\citep{icedit}, BAGEL~\citep{bagel} spanning $\sim$1B--20B parameters and UNet/DiT/MoT families (sizes annotated in Table~\ref{tab:main}). Our backbone reference is the un-tuned FLUX.1~Kontext-dev~\citep{flux_kontext} (12B), which isolates the gain attributable to RL fine-tuning. All methods use identical inference settings (Sec.~\ref{sec:exp_impl}).

\subsection{Implementation Details}
\label{sec:exp_impl}

\textbf{Training.} We fine-tune FLUX.1~Kontext-dev~\citep{flux_kontext} (12B) with MixGRPO~\citep{mixgrpo} as the RL backbone, training for $230$ steps ($\approx$300 GPU-hr on $8\!\times\!\text{A100-80GB}$). The reward is produced by Qwen3-VL-30B-A3B-Instruct~\citep{qwen3} as the audit/verification VLM $\Phi_{\text{V}}$ and Qwen3-30B-A3B-Instruct-2507~\citep{qwen3} as the rubric-aggregation LLM $\Phi_{\text{L}}$, both served via vLLM~\citep{vllm}. Full hyperparameters are in App.~\ref{app:training_settings}.

\textbf{Reward-generation cost.} For a group of 16 candidates, audit, aggregation, and verification took 16.06 seconds and processed 61{,}950 tokens under the hardware and serving configuration in App.~\ref{app:rebuttal_results}. This measures reward generation rather than end-to-end training throughput.

\textbf{Evaluation.} All methods are evaluated under identical inference settings (resolution, denoising steps, CFG, seed; details in App.~\ref{app:training_settings}); metrics follow Sec.~\ref{sec:exp_setup}.

\subsection{Main Results}
\label{sec:exp_main}

\begin{table}[t]
\centering
\caption{Final-checkpoint results on four image-editing benchmarks. We report \textsc{EditScore} (ES) Overall per benchmark, plus the unweighted mean across benchmarks for all three metrics (\textbf{Avg}, mean$_{\pm\text{std}}$ across the four benchmarks). The four benchmarks differ in size by ${\sim}6\times$, so we use unweighted averaging to give equal voice to each benchmark; per-benchmark HPSv2 and ImageReward are in App.~\ref{app:full_main}. Higher is better in all columns; ES $\in [0,10]$, HPS $\in [0,1]$, IR is real-valued. Best per column in \textbf{bold}.}
\label{tab:main}
\setlength{\tabcolsep}{4pt}
\renewcommand{\arraystretch}{1.1}
\small
\resizebox{\linewidth}{!}{%
\begin{tabular}{lccccccc}
\toprule
& \multicolumn{3}{c}{\textbf{Avg}} & \multicolumn{4}{c}{\textbf{ES per benchmark}} \\
\cmidrule(lr){2-4}\cmidrule(lr){5-8}
Method & ES$\uparrow$ & HPS$\uparrow$ & IR$\uparrow$ & MinEval & MagicBr.\ & AnyBench & Emu-Edit \\
\midrule
FLUX.1~Kontext-dev (Base, 12B, 2025)  & 5.21$_{\pm0.34}$ & 0.215$_{\pm0.007}$ & $-$0.67$_{\pm0.14}$ & 4.72 & 5.25 & 5.46 & 5.40 \\
InstructPix2Pix ($\sim$1B, 2022)      & 2.43$_{\pm0.34}$ & 0.211$_{\pm0.009}$ & $-$0.89$_{\pm0.17}$ & 1.93 & 2.50 & 2.63 & 2.65 \\
MagicBrush model ($\sim$1B, 2023)     & 3.53$_{\pm0.63}$ & 0.207$_{\pm0.006}$ & $-$0.90$_{\pm0.07}$ & 3.25 & 4.45 & 3.04 & 3.38 \\
OmniGen2 ($\sim$7B, 2025)             & 5.08$_{\pm0.30}$ & 0.214$_{\pm0.006}$ & $-$0.72$_{\pm0.10}$ & 5.15 & 5.03 & 4.71 & 5.44 \\
UniWorld-v1 ($\sim$20B, 2025)         & 3.43$_{\pm0.26}$ & 0.202$_{\pm0.004}$ & $-$0.92$_{\pm0.05}$ & 3.38 & 3.40 & 3.78 & 3.15 \\
ICEdit ($\sim$12B, 2025)              & 5.15$_{\pm0.57}$ & 0.207$_{\pm0.006}$ & $-$0.81$_{\pm0.08}$ & 4.90 & 5.73 & 4.48 & 5.50 \\
BAGEL ($\sim$14B, 2025)               & 5.80$_{\pm0.43}$ & 0.209$_{\pm0.003}$ & $-$0.73$_{\pm0.10}$ & \textbf{5.98} & \textbf{6.22} & 5.22 & 5.79 \\
\midrule
\rowcolor[gray]{0.92}
\methodname{} (12B, 2026, ours)       & \textbf{5.88$_{\pm0.15}$} & \textbf{0.222$_{\pm0.008}$} & \textbf{$-$0.51$_{\pm0.16}$} & 5.73 & 5.80 & \textbf{6.07} & \textbf{5.91} \\
\bottomrule
\end{tabular}}
\end{table}

\textbf{Cross-benchmark comparison.} \methodname{} achieves the highest unweighted average across ES, HPSv2, and IR (Table~\ref{tab:main}), with gains over the un-tuned base on all four benchmarks. BAGEL leads on MinEval and MagicBrush ES, while \methodname{} leads on AnyBench and Emu-Edit ES; thus the four-benchmark average favors \methodname{} ($5.88$ vs.\ $5.80$). The independent scorers provide complementary quality evidence, while the masked pixel analysis below directly tests off-target preservation.

\textbf{Off-target modification.} On the 600-example masked evaluation, \methodname{} reduces OffTarget-L1 from $0.20366$ (Base) to $0.18659$, an 8.4\% relative reduction. The EditScore(PF, Cons)-GDPO reward control and a separate EditScore-Top1-SFT control obtain $0.18975$ and $0.19952$, respectively (App.~\ref{app:offtarget}). This result directly supports reduced change outside the designated target mask; it does not imply that every pixel inside the mask was correctly edited or that segmentation is error-free.

\textbf{Zero-shot generalization across edit semantics (AnyBench).} \methodname{} is trained on only six atomic edit types, but AnyBench's 18-category taxonomy lets us probe its generalization to unseen edit semantics (Table~\ref{tab:per_edit_type}). The improvements are systemic rather than category-bound: \methodname{} beats the un-tuned base on $16/18$ categories under both EditScore and HPSv2, with the largest gains on tasks where locality is notoriously difficult, e.g., \texttt{movement} ($+1.58$ ES), \texttt{implicit\_chg.}\ ($+1.30$), and \texttt{resize} ($+1.02$). This breadth suggests the dual-channel reward pushes the policy to learn localized image manipulation rather than memorizing specific instruction templates.

\begin{table}[t]
\centering
\caption{Per-edit-type \textsc{EditScore} Overall on AnyBench (50 prompts per edit type). Edit-type names follow AnyBench's taxonomy and may differ in surface form from the six training edit types in App.~\ref{app:data_edit_types} (e.g., \texttt{appearance\_alter} vs.\ \texttt{appearance\_change}, \texttt{action\_chg.}\ vs.\ \texttt{action\_change}). Edit types are columns and methods are rows. Higher is better. The best score in each column is shown in \textbf{bold}, and the second-best score is marked with $^*$. Full results are in App.~\ref{app:per_edit_type}.}
\label{tab:per_edit_type}
\setlength{\tabcolsep}{3pt}
\renewcommand{\arraystretch}{1.0}
\scriptsize
\resizebox{\linewidth}{!}{%
\begin{tabular}{l cccccccccccccccccc}
\toprule
\multirow{2}{*}{Method} & \multicolumn{18}{c}{Edit type} \\
\cmidrule(lr){2-19}
& act. & add & app. & bg. & color & count & impl. & mat. & move & outp. & rel. & rem. & repl. & resize & rot. & style & text & tone \\
\midrule
Base      & 5.76$^*$ & 7.88 & 5.74 & 6.29 & 5.62 & 4.51$^*$ & 5.09$^*$ & 4.18 & 2.51 & 4.71$^*$ & 3.80 & 5.85$^*$ & 5.87 & 3.98$^*$ & 7.10 & 5.37$^*$ & \textbf{8.34} & 5.74 \\
\rowcolor[gray]{0.92}
\methodname{} & \textbf{6.69} & 8.33$^*$ & 6.45$^*$ & \textbf{6.77} & 6.25 & \textbf{5.03} & \textbf{6.39} & 4.57$^*$ & \textbf{4.09} & \textbf{4.89} & \textbf{4.60} & 5.62 & 6.41$^*$ & \textbf{5.00} & 7.47$^*$ & \textbf{5.92} & 8.25$^*$ & 6.53$^*$ \\
IP2P      & 1.48 & 4.62 & 3.02 & 2.89 & 4.99 & 0.42 & 2.72 & 2.61 & 1.95 & 2.48 & 2.35 & 1.46 & 2.50 & 2.19 & 3.02 & 3.42 & 0.08 & 5.06 \\
MagicBr.  & 2.39 & 6.98 & 4.78 & 3.06 & 3.78 & 1.79 & 3.77 & 2.15 & 3.29 & 1.20 & 3.31 & 3.45 & 3.61 & 1.55 & 2.43 & 2.22 & 0.86 & 4.18 \\
OmniGen2  & 2.56 & 7.18 & 5.93 & 6.72$^*$ & 6.46 & 2.32 & 4.42 & 3.98 & 2.16 & 4.62 & 4.02$^*$ & \textbf{6.30} & \textbf{6.74} & 1.61 & 2.70 & 4.55 & 6.78 & 5.71 \\
UniWorld-v1 & 4.38 & 6.09 & 3.88 & 4.20 & 4.76 & 3.48 & 4.93 & 3.49 & 2.95 & 3.59 & 3.60 & 3.84 & 4.46 & 1.90 & 4.51 & 4.14 & 0.96 & 2.88 \\
ICEdit    & 2.40 & 8.15 & 5.98 & 5.74 & 6.63$^*$ & 2.15 & 3.91 & \textbf{4.60} & 2.66 & 2.62 & 3.13 & 4.80 & 5.11 & 2.47 & 2.93 & 5.03 & 5.78 & \textbf{6.60} \\
BAGEL     & 4.57 & \textbf{8.86} & \textbf{6.62} & 5.73 & \textbf{6.79} & 3.67 & 4.74 & 3.78 & 3.28$^*$ & 2.82 & 3.35 & 5.61 & 5.78 & 2.84 & \textbf{8.12} & 4.25 & 7.34 & 5.85 \\
\bottomrule
\end{tabular}}
\end{table}

\subsection{Ablations}
\label{sec:exp_ablation}

\methodname{} combines two critical design choices on top of GDPO~\citep{gdpo}: a \emph{dual-channel} reward $(r^{(1)}, r^{(2)})$ that decouples efficacy from unintended change, and \emph{per-channel group normalization} that turns the two raw reward channels into commensurable advantages $(A^{(1)}, A^{(2)})$ before they are combined. Table~\ref{tab:ablation} enumerates the variants we compare; each isolates a single design factor. All variants share identical backbones, training data, reward models, and hyperparameters, evaluated at the 230-step final checkpoint.

\begin{table}[H]
\centering
\caption{Ablation variants and results. The \textbf{Reward} column gives the reward signal used for RL fine-tuning; variants (2)--(4) toggle one design factor relative to the full method (5). \textbf{Avg} is the unweighted mean across the four benchmarks. Final-checkpoint EditScore Overall ($\uparrow$), HPSv2 ($\uparrow$), ImageReward ($\uparrow$); best per column in \textbf{bold}. Variant (4) reward-hacks to a degenerate noise output and is collapsed to a single line (see text).}
\label{tab:ablation}
\setlength{\tabcolsep}{3pt}
\renewcommand{\arraystretch}{1.05}
\footnotesize
\resizebox{\linewidth}{!}{%
\begin{tabular}{ll ccc ccc ccc ccc ccc}
\toprule
& & \multicolumn{3}{c}{\textbf{Avg}} & \multicolumn{3}{c}{\textbf{MinEval}} & \multicolumn{3}{c}{\textbf{MagicBrush}} & \multicolumn{3}{c}{\textbf{AnyBench}} & \multicolumn{3}{c}{\textbf{Emu-Edit}} \\
\cmidrule(lr){3-5}\cmidrule(lr){6-8}\cmidrule(lr){9-11}\cmidrule(lr){12-14}\cmidrule(lr){15-17}
Variant & Reward & ES$\uparrow$ & HPS$\uparrow$ & IR$\uparrow$ & ES$\uparrow$ & HPS$\uparrow$ & IR$\uparrow$ & ES$\uparrow$ & HPS$\uparrow$ & IR$\uparrow$ & ES$\uparrow$ & HPS$\uparrow$ & IR$\uparrow$ & ES$\uparrow$ & HPS$\uparrow$ & IR$\uparrow$ \\
\midrule
(1)\enspace Base (no RL)                   & \textcolor{gray}{$\varnothing$}                   & 5.21$_{\pm0.34}$ & 0.215$_{\pm0.007}$ & $-$0.67$_{\pm0.14}$ & 4.72 & 0.212 & $-$0.80 & 5.25 & 0.224 & $-$0.51 & 5.46 & 0.215 & $-$0.58 & 5.40 & 0.208 & $-$0.78 \\
(2)\enspace Merged scalar                  & $w_1 r^{(1)} + w_2 r^{(2)}$      & 4.35$_{\pm0.37}$ & 0.211$_{\pm0.006}$ & $-$0.75$_{\pm0.10}$ & 3.89 & 0.209 & $-$0.82 & 4.23 & 0.219 & $-$0.62 & 4.75 & 0.212 & $-$0.72 & 4.52 & 0.205 & $-$0.85 \\
(3)\enspace Efficacy only                  & $r^{(1)}$                                         & 4.09$_{\pm0.45}$ & 0.203$_{\pm0.007}$ & $-$0.87$_{\pm0.17}$ & 3.46 & 0.197 & $-$1.05 & 4.09 & 0.212 & $-$0.65 & 4.49 & 0.204 & $-$0.83 & 4.33 & 0.197 & $-$0.94 \\
(4)\enspace Locality only \textcolor{gray}{[reward-hacked]} & $r^{(2)}$                              & \multicolumn{15}{c}{\textcolor{gray}{ES~$\approx$~0.00\,\enspace HPS~$\approx$~0.011\,\enspace IR~$\approx-2.25$ across all four benchmarks (degenerate noise output)}} \\
\midrule
\rowcolor[gray]{0.92}
(5)\enspace \methodname{} (full, ours)     & $(r^{(1)}, r^{(2)})$ + per-channel adv.           & \textbf{5.88$_{\pm0.15}$} & \textbf{0.222$_{\pm0.008}$} & \textbf{$-$0.51$_{\pm0.16}$} & \textbf{5.73} & \textbf{0.220} & \textbf{$-$0.63} & \textbf{5.80} & \textbf{0.233} & \textbf{$-$0.29} & \textbf{6.07} & \textbf{0.221} & \textbf{$-$0.49} & \textbf{5.91} & \textbf{0.214} & \textbf{$-$0.62} \\
\bottomrule
\end{tabular}}
\end{table}

\textbf{Summary.} Each ablation exposes a distinct failure mode. \emph{Merged scalar (2)}: collapsing the dual channels before group normalization drops EditScore by ${\sim}1.5$, as the higher-variance signal washes out the other, confirming per-channel normalization is essential. \emph{Efficacy only (3)}: without locality pressure, the policy over-edits and corrupts non-target context ($-1.12$ ES vs.\ Base). \emph{Locality only (4)}: by step ${\sim}170$ the policy reward-hacks to high-frequency noise, $|\mathcal{O}_i|\!\to\!0$, and $r^{(2)}$ saturates even though every instruction is unimplemented. The dual-channel design (5) closes both shortcuts: $r^{(1)}$ anchors semantic validity so neither over-editing nor empty outputs is rewarded.

\textbf{Reward construction and optimization controls.} With GDPO and the training configuration fixed, EditScore(PF, Cons) rewards yield $3.777$ average ES across the four benchmarks, versus $5.880$ for ARRO rewards. Removing aggregation while retaining verification reduces the four-benchmark average ES from $5.880$ to $4.961$, supporting the value of a shared, consolidated rubric. In a separate MinEval-only comparison, raw audit counts yield $3.2779$ ES, ARRO Top-10 SFT yields $4.6289$, and ARRO-GDPO yields $5.6690$; these values are not four-benchmark averages. In a controlled CLEVR diagnostic with all requested recolorings implemented, ARRO assigns zero unintended changes for 5--50 requested edits, whereas EditScore Consistency varies despite zero ground-truth unintended changes (App.~\ref{app:rebuttal_results}). Detailed protocols and metrics are in App.~\ref{app:rebuttal_results}.

\textbf{Human validation and backbone transfer.} Two blinded annotators checked 60 candidate groups (960 candidates) and 565 final issues: they jointly confirmed correct merging for $560/565$ issues, visual support for $553/565$, and no evident omitted issue in $58/60$ groups. In a separate blinded pairwise study, ARRO wins more comparisons than it loses against EditScore(PF, Cons)-GDPO (Table~\ref{tab:human_pairwise}). Applying the same reward pipeline to OmniGen2 with its FlowGRPO optimizer improves ES on all three unseen benchmarks (Table~\ref{tab:omnigen_transfer_main}) and wins more human comparisons than the original OmniGen2 (Table~\ref{tab:human_pairwise}). Study protocols and full transfer metrics are in App.~\ref{app:human_eval} and App.~\ref{app:omnigen_transfer}.

\begin{table}[t]
\centering
\caption{Blinded pairwise judgments on 60 MinEval examples by ten evaluators (600 judgments per comparison). Win, tie, and loss are from the ARRO variant's perspective; judgments share 60 underlying image pairs.}
\label{tab:human_pairwise}
\small
\begin{tabular}{lccc}
\toprule
Comparison & Win & Tie & Loss \\
\midrule
ARRO vs. EditScore(PF, Cons)-GDPO & 366/600 (61.0\%) & 94/600 (15.7\%) & 140/600 (23.3\%) \\
OmniGen2 + ARRO vs. OmniGen2 & 321/600 (53.5\%) & 107/600 (17.8\%) & 172/600 (28.7\%) \\
\bottomrule
\end{tabular}
\end{table}

\begin{table}[t]
\centering
\caption{OmniGen2 transfer on three unseen benchmarks. EditScore (ES) is higher-is-better; the average is unweighted. Full HPSv2 and ImageReward results are in App.~\ref{app:omnigen_transfer}.}
\label{tab:omnigen_transfer_main}
\small
\begin{tabular}{lccc}
\toprule
Benchmark & OmniGen2 & OmniGen2 + ARRO & ES gain \\
\midrule
MagicBrush & 5.03 & 5.58 & +0.55 \\
AnyBench & 4.71 & 5.22 & +0.51 \\
Emu-Edit & 5.44 & 5.99 & +0.55 \\
\midrule
Average & 5.06 & 5.59 & +0.53 \\
\bottomrule
\end{tabular}
\end{table}

\section{Conclusion}
\label{sec:conclusion}

We recast \emph{minimal change} for instruction-based image editing as an outcome-level RL objective with separate rewards for unimplemented and unintended changes. On FLUX.1~Kontext-dev, \methodname{} improves average EditScore from $5.21$ to $5.88$ across four benchmarks and reduces masked off-target pixel change by 8.4\%. Reward and SFT controls, human evaluations, and OmniGen2 transfer support the value of structured reward construction and direct optimization.

\paragraph{Limitations and broader impacts.}\label{sec:limitations} Reward quality is bounded by the VLM's perception, especially for subtle differences, ambiguous or compositional instructions, and candidate-specific issue assignment. An imperfect judge can introduce systematic errors or be exploited during RL; the expert study validates the evaluated configuration rather than all possible auditors. OffTarget-L1 depends on segmentation and captures pixel change rather than full semantic correctness. We demonstrate a second rectified-flow editor, OmniGen2, but do not establish transfer to diffusion backbones or quantify multi-seed variability. More controllable editing could also facilitate deceptive image manipulation; deployment requires provenance and misuse safeguards beyond the reward mechanism studied here.

\section{Acknowledgments}
Tongliang Liu is partially supported by the following Australian Research Council projects: DP260102466, FT220100318, DP220102121, LP220100527, LP220200949. 
Yu Yao is partially supported by the following Australian Research Council projects: DE260101993, DP260102466.

\bibliography{references}
\bibliographystyle{plainnat}

\appendix

\section{Related Work (Detailed)}

\label{app:ref}

\paragraph{Instruction-based image editing: pretraining.} The dominant paradigm for instruction-based image editing trains a conditional generative model on large-scale (source, instruction, edit) triplets. InstructPix2Pix~\citep{instructpix2pix} pioneered this direction by synthesizing paired training data from a language model and a text-to-image model, enabling a forward-pass diffusion editor to follow free-form instructions without per-example optimization. Subsequent work scaled this recipe along two axes: stronger language understanding through multimodal LLM backbones~\citep{smartedit,instructany2pix}, and stronger visual backbones through diffusion-transformer architectures and rectified flow, most notably FLUX.1~Kontext~\citep{flux_kontext} and Qwen-Image-Edit~\citep{qwen_image_edit}, which explicitly target locality and minimal background drift as a design goal. Across this line of work, locality is treated as an emergent property of data quality, conditioning design, and model capacity; it is not optimized for directly. As reported by recent benchmarks~\citep{compbench,gie_bench,gedit_bench_v2}, even the strongest pretrained editors still leak edits into the background, shift identity, or introduce unintended attribute changes, motivating a post-training stage that targets these failures explicitly.

\paragraph{Instruction-based image editing: post-training.}
A second line of work post-trains pretrained editors using preference
or reward feedback. Diffusion-DPO~\citep{diffusion_dpo} adapts direct
preference optimization to diffusion models, with follow-up work
extending it to richer or per-region preference
signals~\citep{rpo,personalized_dpo,spie}. On the online-RL side,
Flow-GRPO~\citep{flowgrpo} and DanceGRPO~\citep{dancegrpo} bring
group-relative policy optimization to flow-matching generators, and
GDPO~\citep{gdpo} extends this to multiple rewards. For the editing
setting specifically, EditScore~\citep{editscore},
OneReward~\citep{onereward}, RewardDance~\citep{rewarddance},
SpatialReward~\citep{spatialreward}, and
ThinkRL-Edit~\citep{thinkrl_edit_2026} train dedicated reward models
to drive online RL fine-tuning.
Our work differs from this line on two fronts. \emph{First, human
preference is not minimal change.} Prior rewards fit human preference
data, which mixes many factors at once: visual quality, instruction
adherence, aesthetic appeal, and overall plausibility. Whether the
edit altered things outside the intended scope is only one of these
factors, often the least salient one to a human rater glancing at the
result. A reward trained to predict human preference therefore has no
reason to penalize unintended changes specifically, especially when
the edit otherwise looks plausible. We target minimal change directly
instead. \emph{Second, prior rewards are not scalable.} Each of
EditScore, OneReward, RewardDance, SpatialReward, and ThinkRL-Edit is
a learned reward model, requiring tens to hundreds of thousands of
annotated preference pairs and a training pipeline of its own. Our
reward requires no preference data and no reward-model training: an
off-the-shelf VLM is prompted to audit each edit, and the two reward
components for minimal change are read off directly from its output.

\paragraph{Minimal change in causal representation learning.} The principle that an intervention should perturb only the targeted mechanism originates in causal inference, where it underpins Pearl's $do$-calculus and the structural-causal-model account of counterfactuals~\citep{pearl_causality,peters_elements_2017}. In representation learning, this principle is formalized as the \emph{sparse mechanism shift} hypothesis~\citep{scholkopf_toward_2021,perry_smshift_2022,lopez_smshift_2022}: distribution shifts across environments correspond to sparse changes in a small number of causal mechanisms, and exploiting this sparsity yields identifiability of the latent causal factors~\citep{lachapelle_partial_2024, xu_sparsity_2024}. Within deep generative models, CausalVAE~\citep{causalvae}, DEAR~\citep{dear}, SCM-VAE~\citep{scmvae}, and more recent flow-prior variants~\citep{flexcausal_2026,causaldiffae} operationalize minimal change by structuring the latent space around a learned causal graph and pairing it with weak label or counterfactual supervision, so that interventions can be directed at individual latent factors. A parallel, training-free line of work in StyleGAN-era image editing makes the same intuition explicit by adding $L_1$ or sparse-autoencoder constraints to the latent edit at inversion time~\citep{ppe_sparse_2025,saedit_2025}. Our work follows the same conceptual goal but disagrees on where the constraint should live: we argue that minimal change is fundamentally a property of the \emph{output} of an intervention, and that enforcing it on a latent that is then passed through a deeply nonlinear decoder is a brittle proxy.    

\paragraph{Minimal change for image generation and editing.} The same principle has been brought to image-level tasks via counterfactual image generation under structural causal models~\citep{causaldiffae,diffusion_counterfactual_abduction,medical_causal_disentangle,counterscene} and StyleGAN-era disentangled face editing with axis-aligned latents or training-free $L_1$/layer-masking constraints~\citep{ppe_sparse_2025,saedit_2025}. These methods deliver strong locality on narrow domains with attribute labels or counterfactual pairs, but inherit the same three obstacles we identify in Sec.~\ref{sec:intro} when scaled to free-form instructions; we sidestep all three by measuring minimal change on the generated image and optimizing it end-to-end via an agentic reward.   

\section{Optimization Details}
\label{app:gdpo_full}

This section spells out the GDPO optimizer and our dual-channel advantage in full, complementing the abbreviated description in Sec.~\ref{sec:prelim} and Sec.~\ref{sec:opt}.

\paragraph{Rectified-flow training objective.}
We restate Eq.~\eqref{eq:rf} for reference. FLUX.1~Kontext is pretrained as a rectified-flow transformer~\citep{rectified_flow,flux_kontext}. With $\bm{z}_0\!\sim\!p_{\text{data}}$, $\bm{\epsilon}\!\sim\!\mathcal{N}(0,I)$, $t\!\sim\!\mathcal{U}[0,1]$, and source-image latent $\bm{z}_{\text{src}}$ concatenated with the noisy latent $\bm{z}_t=(1{-}t)\bm{z}_0+t\bm{\epsilon}$ along the sequence axis, the velocity field $\bm{v}_\theta$ is trained with
\begin{equation}
\mathcal{L}_{\text{RF}} = \mathbb{E}_{t,\bm{z}_0,\bm{\epsilon}}\,\big\|\bm{v}_\theta\big([\bm{z}_t;\bm{z}_{\text{src}}],\,t,\,c\big) - (\bm{\epsilon}-\bm{z}_0)\big\|_2^2.
\label{eq:flow}
\end{equation}
The pretraining loss~\eqref{eq:flow} is not used during our RL fine-tuning; we include it only to fix notation.

\paragraph{GDPO advantage.}
For an input pair $(\bm{x}_{\text{src}},c)\!\sim\!\mathcal{D}$, GDPO~\citep{gdpo} samples a group $\mathcal{G}=\{\bm{x}_1,\dots,\bm{x}_N\}$ of $N$ candidates from the current policy $\pi_{\theta_{\text{old}}}(\,\cdot\!\mid\!\bm{x}_{\text{src}},c)$ and scores each candidate by $n$ reward channels $\{r^{(k)}_i\}_{k=1}^{n}$. The advantage used in policy optimization is constructed in three steps. \emph{Step 1 (per-reward group $z$-score).} Each channel is normalized within the group,
\begin{equation}
A^{(k)}_i \;=\; \frac{r^{(k)}_i - \mu^{(k)}_\mathcal{G}}{\sigma^{(k)}_\mathcal{G}+\varepsilon_0},
\qquad
\mu^{(k)}_\mathcal{G} = \tfrac{1}{N}\!\sum_{j=1}^{N} r^{(k)}_j,
\quad
\sigma^{(k)}_\mathcal{G} = \sqrt{\tfrac{1}{N}\!\sum_{j=1}^{N}(r^{(k)}_j-\mu^{(k)}_\mathcal{G})^2},
\label{eq:gdpo_adv}
\end{equation}
where $\varepsilon_0$ is a small numerical-stability offset. \emph{Step 2 (weighted sum across channels).} The per-channel advantages are aggregated into a group-level advantage
\begin{equation}
A^{\text{sum}}_i \;=\; \sum_{k=1}^{n} w_k\, A^{(k)}_i, \qquad w_k > 0.
\label{eq:gdpo_sum}
\end{equation}
\emph{Step 3 (batch-level $z$-score).} Finally, $A^{\text{sum}}_i$ is renormalized across the training batch $\mathcal{B}$ (all candidates from all groups in $\mathcal{B}$) to keep its scale stable as the number of channels $n$ grows:
\begin{equation}
\hat{A}_i \;=\; \frac{A^{\text{sum}}_i - \mu^{\text{sum}}_\mathcal{B}}{\sigma^{\text{sum}}_\mathcal{B}+\varepsilon_0},
\qquad
\mu^{\text{sum}}_\mathcal{B} = \mathrm{mean}_{j\in\mathcal{B}}\,A^{\text{sum}}_j,
\qquad
\sigma^{\text{sum}}_\mathcal{B} = \mathrm{std}_{j\in\mathcal{B}}\,A^{\text{sum}}_j.
\label{eq:gdpo_batchnorm}
\end{equation}
Step 3 is an affine, batch-wide transform: it preserves within-group rankings (and therefore the relative ordering used by PPO) but stabilizes the magnitude that the clip threshold $\epsilon_{\text{clip}}$ sees.

\paragraph{MixGRPO sampling.}
Sampling from $\pi_\theta$ uses an SDE discretization of the rectified flow~\citep{mixgrpo} over $T$ denoising steps indexed $t\!=\!T,\dots,1$ (a discrete index, distinct from the continuous flow time of Eq.~\eqref{eq:flow}), inducing transition kernels $\pi_\theta(\bm{x}_{i,t-1}\!\mid\!\bm{x}_{i,t})$ along each candidate's trajectory. With the per-step importance ratio
\begin{equation}
\rho_{i,t}(\theta)\;=\;\frac{\pi_\theta(\bm{x}_{i,t-1}\!\mid\!\bm{x}_{i,t})}{\pi_{\theta_{\text{old}}}(\bm{x}_{i,t-1}\!\mid\!\bm{x}_{i,t})},
\end{equation}
the policy is updated by the PPO-clip objective
\begin{equation}
\mathcal{J}_{\text{GDPO}}(\theta) \;=\; \mathbb{E}_{i,t}\Big[\min\!\big(\rho_{i,t}(\theta)\,\hat{A}_i,\;\mathrm{clip}(\rho_{i,t}(\theta),\,1{-}\epsilon_{\text{clip}},\,1{+}\epsilon_{\text{clip}})\,\hat{A}_i\big)\Big],
\label{eq:gdpo_obj}
\end{equation}
where $\hat{A}_i$ is the batch-normalized advantage from~\eqref{eq:gdpo_batchnorm}, and the clip range $\epsilon_{\text{clip}}$ is distinct from the stability constant $\varepsilon_0$ in~\eqref{eq:gdpo_adv}/\eqref{eq:gdpo_batchnorm} and from the noise variable $\bm{\epsilon}$ in~\eqref{eq:flow}.

\paragraph{Our dual-channel advantage.}
We instantiate the GDPO pipeline above with $n\!=\!2$. The per-channel advantages $A^{(1)}_i, A^{(2)}_i$ are obtained from~\eqref{eq:gdpo_adv} with $r^{(k)}_i$ being the unimplemented and unintended reward counts of Eq.~\eqref{eq:rewards}, the group-level advantage $A^{\text{sum}}_i = w_1 A^{(1)}_i + w_2 A^{(2)}_i$ follows~\eqref{eq:gdpo_sum} (cf.\ Eq.~\eqref{eq:our_advantage_sum}), and the batch-normalized advantage $\hat{A}_i$ is given by~\eqref{eq:gdpo_batchnorm} (cf.\ Eq.~\eqref{eq:our_advantage}). Plugging $\hat{A}_i$ into~\eqref{eq:gdpo_obj} yields the final training objective
\begin{equation}
\mathcal{J}_{\text{ours}}(\theta) \;=\; \mathbb{E}_{(\bm{x}_{\text{src}},c)\sim\mathcal{D}}\;\mathbb{E}_{\mathcal{G}\sim\pi_{\theta_{\text{old}}}}\;\mathbb{E}_{i,t}\!\Big[\min\!\big(\rho_{i,t}(\theta)\,\hat{A}_i,\;\mathrm{clip}\!\big(\rho_{i,t}(\theta),\,1{-}\epsilon_{\text{clip}},\,1{+}\epsilon_{\text{clip}}\big)\,\hat{A}_i\big)\Big],
\label{eq:final_obj}
\end{equation}
where the outer expectation is over training pairs from $\mathcal{D}$, the middle expectation over groups sampled from $\pi_{\theta_{\text{old}}}$, and the inner expectation over candidate index $i$ and denoising step $t$.

\section{Additional Experimental Results}
\label{app:extra_results}

\subsection{Additional Validation}
\label{app:rebuttal_results}

\paragraph{Off-target pixel modification.}
\label{app:offtarget}
We randomly selected 600 examples from the evaluation benchmarks. For each source image $\bm{x}_{\rm src}$ and instruction, SAM~2~\citep{sam2} was used to segment the target object and produce a mask $M$ covering the region expected to change. For edited image $\bm{x}_{\rm edit}$, we measure the mean absolute difference outside that mask:
\begin{equation}
\operatorname{OffTarget\text{-}L1}=
\frac{\left\|(1-M)\odot(\bm{x}_{\rm edit}-\bm{x}_{\rm src})\right\|_1}
{\left\|1-M\right\|_1}.
\end{equation}
All four methods in Table~\ref{tab:offtarget} use the same mask protocol. EditScore(PF, Cons)-GDPO replaces ARRO's two issue-count rewards with EditScore Prompt-Following and Consistency under fixed GDPO (Table~\ref{tab:editscore_reward}). EditScore-Top1-SFT instead fine-tunes on the highest-EditScore candidate in each source--instruction group; it is distinct from Top-10 SFT, which selects the highest-scoring 10\% of MinEval candidates (Table~\ref{tab:reward_sft}). OffTarget-L1 also depends on mask quality and image alignment and does not measure semantic correctness.
\begin{table}[H]
\centering
\caption{Off-target pixel change on 600 randomly selected evaluation examples. Relative improvement is computed against the base editor. Lower OffTarget-L1 is better.}
\label{tab:offtarget}
\begin{tabular}{lcc}
\toprule
Method & OffTarget-L1 $\downarrow$ & Improvement vs. Base \\
\midrule
FLUX.1 Kontext-dev (Base) & 0.20366 & --- \\
EditScore-Top1-SFT & 0.19952 & 2.0\% \\
EditScore(PF, Cons)-GDPO & 0.18975 & 6.8\% \\
ARRO & \textbf{0.18659} & \textbf{8.4\%} \\
\bottomrule
\end{tabular}
\end{table}

\paragraph{Reward construction and offline selection.}
The raw-audit variant directly uses initial open-ended issue counts, without aggregation or rubric-anchored verification. The Top-10 SFT variants select the highest-reward 10\% of MinEval candidates under either ARRO or EditScore reward and use them as targets for LoRA fine-tuning. ARRO-GDPO instead uses within-group relative rewards across candidates. Table~\ref{tab:reward_sft} reports a separate MinEval-only comparison on the 600-example split: its full-method ES is $5.6690$, while the four-benchmark experiment in Table~\ref{tab:main} reports $5.73$ on MinEval. The four-benchmark mean uses the MinEval result from that experiment. The no-aggregation variant retains verification without consolidating semantically overlapping issues; its four-benchmark means are ES $4.961$, HPSv2 $0.2101$, and ImageReward $-0.7672$, compared with $5.880$, $0.2220$, and $-0.5100$ for full ARRO. Raw-audit results are only available on MinEval, so we do not present them as a four-benchmark mean.
\begin{table}[H]
\centering
\caption{Reward-pipeline and Top-10 SFT controls on the same 600-example MinEval evaluation. All methods use the same EditScore, HPSv2, and ImageReward protocols.}
\label{tab:reward_sft}
\setlength{\tabcolsep}{4pt}
\small
\begin{tabular}{lcccccc}
\toprule
Method & PF $\uparrow$ & Cons. $\uparrow$ & PQ $\uparrow$ & ES $\uparrow$ & HPSv2 $\uparrow$ & IR $\uparrow$ \\
\midrule
Base & 5.9920 & 8.1573 & 6.3393 & 4.7237 & 0.2120 & $-0.7991$ \\
Raw audit counts & 5.0843 & 7.0832 & 4.0630 & 3.2779 & 0.1911 & $-1.0830$ \\
EditScore Top-10 SFT & 5.7700 & 8.2827 & 6.2713 & 4.5529 & 0.2091 & $-0.8467$ \\
ARRO Top-10 SFT & 5.7893 & 8.3240 & 6.3933 & 4.6289 & 0.2082 & $-0.8782$ \\
ARRO-GDPO & \textbf{7.1113} & 8.1700 & \textbf{6.5693} & \textbf{5.6690} & \textbf{0.2196} & \textbf{$-0.6309$} \\
\bottomrule
\end{tabular}
\end{table}

\paragraph{EditScore rewards under fixed GDPO.}
To test whether the gain is merely due to GDPO or reward objectives that resemble EditScore, we replace ARRO's two verified error-count signals with EditScore Prompt-Following and Consistency scores while retaining the GDPO configuration and training examples. Table~\ref{tab:editscore_reward} shows that ARRO's structured rewards outperform this control on all four benchmarks. These are evaluation ES values; the EditScore sub-scores used as training rewards in the control are distinct from the reported overall ES.
\begin{table}[H]
\centering
\caption{EditScore overall under fixed GDPO training with different reward signals. Average is unweighted across four benchmarks.}
\label{tab:editscore_reward}
\begin{tabular}{lccccc}
\toprule
Reward & MinEval & MagicBrush & AnyBench & Emu-Edit & Average \\
\midrule
EditScore(PF, Cons) & 3.566 & 3.233 & 4.379 & 3.932 & 3.777 \\
ARRO verified issue counts & \textbf{5.730} & \textbf{5.800} & \textbf{6.070} & \textbf{5.910} & \textbf{5.880} \\
\bottomrule
\end{tabular}
\end{table}
On the three unseen benchmarks alone (MagicBrush, AnyBench, and Emu-Edit), the corresponding ES averages are $3.848$ for EditScore(PF, Cons)-GDPO and $5.927$ for ARRO. These three-benchmark means exclude MinEval and differ from the four-benchmark averages in Table~\ref{tab:editscore_reward}.

\paragraph{Controlled test of intended versus unintended change.}
In a CLEVR-based~\citep{clevr} diagnostic, each candidate implements all object recolorings specified by its instruction, so the ground-truth unintended-change count is zero. The number of requested recolorings increases from 5 to 50, with five samples per condition. ARRO uses its training-time Audit--Aggregation--Verification procedure; EditScore Consistency uses the public Qwen3-VL-8B evaluator and its official LoRA adapter with three inference runs. The reported means in Table~\ref{tab:clevr_cons} show that the verified ARRO count remains zero across these controlled conditions, while EditScore Consistency varies despite no off-target change. This diagnostic probes the two evaluators' treatment of intended edits; it does not establish behavior on all natural images.
\begin{table}[H]
\centering
\caption{CLEVR diagnostic: every specified modification is implemented and the ground-truth unintended-change count is zero. Each column reports a five-sample mean.}
\label{tab:clevr_cons}
\setlength{\tabcolsep}{3pt}
\small
\begin{tabular}{lcccccccccc}
\toprule
Requested modifications & 5 & 10 & 15 & 20 & 25 & 30 & 35 & 40 & 45 & 50 \\
\midrule
ARRO unintended count & 0 & 0 & 0 & 0 & 0 & 0 & 0 & 0 & 0 & 0 \\
EditScore Consistency & 9.200 & 8.333 & 8.867 & 8.467 & 8.733 & 8.667 & 8.533 & 9.240 & 8.867 & 9.000 \\
\bottomrule
\end{tabular}
\end{table}

\paragraph{Evaluator scale and reward-generation cost.}
For a candidate group of size $N$, the reward pipeline makes $N$ VLM audit calls, one LLM aggregation call, and $N$ VLM verification calls, for $2N+1$ inference passes. With $N=16$, Qwen3-VL-30B-A3B-Instruct and Qwen3-30B-A3B-Instruct-2507 were served on separate NVIDIA A100-SXM4-80GB GPUs at maximum concurrency 128. The complete reward-generation stage took 16.06 seconds per group and processed 61,950 tokens (54,720 prompt and 7,230 completion tokens). Candidate-level audit and verification requests can run concurrently; this measured latency is for the stated hardware and server configuration, not an end-to-end training throughput result.

We also tested smaller evaluator pairs on a separate deterministic CLEVR~\citep{clevr} diagnostic: five scenes, each under 0, 1, 2, or 3 visually salient off-target modifications, giving 20 outputs and five samples per ground-truth count. All five requested object changes were implemented. The VLM performs audit and verification, and the paired LLM performs aggregation, with identical images, prompts, decoding settings, and pipeline. Table~\ref{tab:evaluator_scale} reports the mean predicted count by ground-truth condition, sample-level MAE, and exact-count accuracy. These results support a cost--accuracy trade-off for this controlled count task, without measuring full editing-model training at each evaluator scale.
\begin{table}[H]
\centering
\caption{Evaluator-size diagnostic on 20 CLEVR outputs. GT columns show the average predicted unintended-change count among five examples with that ground-truth count.}
\label{tab:evaluator_scale}
\begin{tabular}{lcccccc}
\toprule
VLM + LLM & GT=0 & GT=1 & GT=2 & GT=3 & MAE $\downarrow$ & Exact acc. $\uparrow$ \\
\midrule
Qwen3 4B + 4B & 0.80 & 1.40 & 1.80 & 3.40 & 0.75 & 0.40 \\
Qwen3 8B + 8B & 0.80 & 1.00 & 2.00 & 2.80 & 0.55 & 0.50 \\
Qwen3 32B + 30B & 0.40 & 1.20 & 1.80 & 3.00 & \textbf{0.30} & \textbf{0.70} \\
\bottomrule
\end{tabular}
\end{table}

\paragraph{Motivation and scope of MinEval.}
Each MinEval instruction targets one atomic editing objective and allows multiple valid outputs. For each source--instruction pair, grouped candidate edits and separate intended- and unintended-change assessments provide graded supervision without one ground-truth edited image. This structure supports the RL comparison studied here and candidate filtering, as illustrated by the Top-10 SFT controls. Uses for disentangled representation learning remain prospective rather than experimentally established here.

\subsection{Human Evaluation}
\label{app:human_eval}
\paragraph{Expert validation of issue judgments.}
We sampled 60 MinEval groups balanced across addition, removal, replacement, movement, action change, and appearance change. Each group contains a shared source image and instruction with 16 edited candidates, yielding 960 candidate images and 565 final aggregated issues. Candidate identities and presentation order were randomized. Two independent second-year computer-science master's students, not involved in this work, reviewed the source, instruction, candidate images, raw audit issues, and verified issues without model identities, reward values, or reward-generation procedures. They independently checked whether repeated issues were merged, whether final issues were supported by the source and edited images, and whether an evident failure was missing. Table~\ref{tab:expert_validation} reports conservative counts jointly confirmed by both annotators; these are issue- and group-level correctness checks, not inter-rater agreement coefficients. Candidate assignment remained a source of error and is not quantified here.
\begin{table}[H]
\centering
\caption{Expert validation of the current reward pipeline. Fractions use issues for the first two rows and candidate groups for the third.}
\label{tab:expert_validation}
\begin{tabular}{lcc}
\toprule
Check & No error confirmed by both & Error confirmed by both \\
\midrule
Repeated issues merged correctly & 560/565 (99.1\%) & 5/565 (0.9\%) \\
Final issue visually supported & 553/565 (97.9\%) & 12/565 (2.1\%) \\
No evident issue omitted & 58/60 (96.7\%) & 2/60 (3.3\%) \\
\bottomrule
\end{tabular}
\end{table}

\paragraph{Blinded pairwise preference.}
We sampled 60 MinEval examples balanced across the same six edit types. Ten independent second-year computer-science master's students, separate from the issue annotators and not involved in this work, each compared the source image, instruction, and two model outputs for all 60 examples. Model identities and metric scores were hidden; left/right presentation was randomized. Evaluators chose the better edit or a tie. The 600 judgments per comparison in Table~\ref{tab:human_pairwise} share 60 underlying examples, so they should not be interpreted as 600 independent image pairs. The prompt asked for overall editing quality rather than a separate locality-only preference.
\subsection{Transfer to OmniGen2}
\label{app:omnigen_transfer}
To assess transfer beyond FLUX.1 Kontext-dev, we trained OmniGen2 with the same ARRO Audit--Aggregation--Verification reward construction and rejection-sampling strategy. OmniGen2 used its official training script and recommended configuration with FlowGRPO; thus the generator and RL implementation both differ from the FLUX.1 Kontext-dev/MixGRPO experiment. The three unseen benchmark results in Table~\ref{tab:omnigen_transfer} consistently improve under this adaptation. The same 60 MinEval examples, evaluators, and blinded protocol used in the pairwise study above give the human result in Table~\ref{tab:human_pairwise}. This demonstrates transfer to a second rectified-flow editor, without claiming diffusion-backbone validation.
\begin{table}[H]
\centering
\caption{OmniGen2 transfer on three unseen benchmarks. Average is the unweighted mean of benchmark scores.}
\label{tab:omnigen_transfer}
\setlength{\tabcolsep}{4pt}
\small
\begin{tabular}{llccc}
\toprule
Benchmark & Model & ES $\uparrow$ & HPSv2 $\uparrow$ & ImageReward $\uparrow$ \\
\midrule
MagicBrush & OmniGen2 & 5.03 & 0.221 & $-0.59$ \\
 & OmniGen2 + ARRO & \textbf{5.58} & \textbf{0.228} & \textbf{$-0.33$} \\
AnyBench & OmniGen2 & 4.71 & 0.215 & $-0.68$ \\
 & OmniGen2 + ARRO & \textbf{5.22} & \textbf{0.221} & \textbf{$-0.58$} \\
Emu-Edit & OmniGen2 & 5.44 & 0.207 & $-0.77$ \\
 & OmniGen2 + ARRO & \textbf{5.99} & \textbf{0.213} & \textbf{$-0.60$} \\
\midrule
Average & OmniGen2 & 5.06 & 0.214 & $-0.68$ \\
 & OmniGen2 + ARRO & \textbf{5.59} & \textbf{0.221} & \textbf{$-0.50$} \\
\bottomrule
\end{tabular}
\end{table}

\subsection{EditScore Sub-Scores}
\label{app:metric_subscores}

The EditScore evaluator~\citep{editscore} produces three sub-scores in addition to the Overall score reported in the main paper:
\begin{itemize}[leftmargin=1.2em,topsep=2pt,itemsep=1pt]
  \item \textbf{EditScore-PF (PF)} --- prompt following: did the requested edit happen? Range $[0,10]$, $\uparrow$.
  \item \textbf{EditScore-Cons (Cons)} --- consistency: are non-target regions preserved? Range $[0,10]$, $\uparrow$.
  \item \textbf{EditScore-PQ (PQ)} --- perception quality: is the output clean and artifact-free? Range $[0,10]$, $\uparrow$.
\end{itemize}
The Overall score reported in Tables~\ref{tab:main} and~\ref{tab:ablation} is a joint judgment combining these three axes. Table~\ref{tab:editscore_subscores_app} reports the per-benchmark PF/Cons/PQ breakdown for all methods, complementing the Overall numbers in Table~\ref{tab:main}. Two observations relevant to the main-paper discussion: \textbf{(i)}~\methodname{} delivers a large gain on prompt-following over the base model ($+0.75$ averaged over the four benchmarks: $6.60\!\to\!7.35$), while its Consistency stays essentially flat ($8.36\!\to\!8.31$, $-0.05$ on average), confirming that the Overall improvement is driven by better instruction execution rather than by larger, more disruptive edits. \textbf{(ii)}~Among the baselines, the largest jumps in Overall are typically explained by PF, not by Cons; methods such as InstructPix2Pix and the MagicBrush model achieve middling Cons by under-editing (low PF), illustrating why the joint Overall metric is the appropriate headline number.

\begin{table}[t]
\centering
\caption{Per-benchmark \textsc{EditScore} sub-scores (PF, Cons, PQ) for all methods, complementing the Overall numbers in Table~\ref{tab:main}. All sub-scores lie in $[0,10]$, $\uparrow$. \textbf{Avg} is the unweighted mean across the four benchmarks. Best per column in \textbf{bold}.}
\label{tab:editscore_subscores_app}
\setlength{\tabcolsep}{3pt}
\renewcommand{\arraystretch}{1.05}
\footnotesize
\resizebox{\linewidth}{!}{%
\begin{tabular}{l ccc ccc ccc ccc ccc}
\toprule
& \multicolumn{3}{c}{\textbf{MinEval}} & \multicolumn{3}{c}{\textbf{MagicBrush}} & \multicolumn{3}{c}{\textbf{AnyBench}} & \multicolumn{3}{c}{\textbf{Emu-Edit}} & \multicolumn{3}{c}{\textbf{Avg}} \\
\cmidrule(lr){2-4}\cmidrule(lr){5-7}\cmidrule(lr){8-10}\cmidrule(lr){11-13}\cmidrule(lr){14-16}
Method & PF & Cons & PQ & PF & Cons & PQ & PF & Cons & PQ & PF & Cons & PQ & PF & Cons & PQ \\
\midrule
FLUX.1~Kontext-dev (Base) & 6.00 & 8.12 & 6.39 & 6.74 & 8.50 & 6.57 & 6.64 & 8.48 & 6.70 & 7.03 & 8.33 & 6.15 & 6.60 & 8.36 & 6.45 \\
InstructPix2Pix           & 2.31 & 6.13 & 6.24 & 3.24 & 6.65 & 6.20 & 3.34 & 6.98 & 6.39 & 3.34 & 6.60 & 6.06 & 3.06 & 6.59 & 6.22 \\
MagicBrush model          & 4.05 & 5.94 & 4.82 & 5.63 & 7.22 & 5.29 & 4.44 & 5.63 & 4.37 & 4.73 & 6.13 & 5.02 & 4.71 & 6.23 & 4.87 \\
OmniGen2                  & 6.03 & 8.77 & \textbf{7.13} & 6.11 & \textbf{9.04} & 7.22 & 6.03 & 8.46 & 6.76 & 6.78 & \textbf{8.54} & 6.49 & 6.24 & \textbf{8.70} & 6.90 \\
UniWorld-v1               & 5.08 & 5.56 & 3.95 & 5.12 & 6.18 & 3.97 & 5.27 & 5.97 & 4.67 & 5.14 & 5.64 & 3.80 & 5.15 & 5.84 & 4.10 \\
ICEdit                    & 5.50 & \textbf{8.78} & \textbf{7.56} & 6.50 & 8.74 & \textbf{7.33} & 5.05 & 8.15 & \textbf{7.12} & 6.21 & 8.37 & \textbf{6.93} & 5.81 & 8.51 & \textbf{7.24} \\
BAGEL                     & \textbf{7.81} & 8.19 & 6.04 & \textbf{8.13} & 8.56 & 6.18 & 6.80 & 8.01 & 5.95 & \textbf{7.91} & 7.78 & 5.79 & \textbf{7.67} & 8.14 & 5.99 \\
\midrule
\rowcolor[gray]{0.92}
\methodname{} (ours)      & 7.14 & 8.19 & 6.60 & 7.48 & 8.31 & 6.56 & \textbf{7.28} & \textbf{8.51} & 6.90 & 7.49 & 8.23 & 6.45 & 7.35 & 8.31 & 6.63 \\
\bottomrule
\end{tabular}}
\end{table}

\subsection{Per-Benchmark HPSv2 and ImageReward}
\label{app:full_main}

Table~\ref{tab:main_full_app} reports the per-benchmark HPSv2 and ImageReward values that complement the per-benchmark EditScore Overall in Table~\ref{tab:main}. \methodname{} is highest on HPSv2 on every benchmark (tied with InstructPix2Pix on MinEval) and highest on ImageReward on every benchmark.

\begin{table}[t]
\centering
\caption{Per-benchmark HPSv2 (HPS) and ImageReward (IR) for all methods. Best per column in \textbf{bold}. EditScore Overall per benchmark is in Table~\ref{tab:main}.}
\label{tab:main_full_app}
\setlength{\tabcolsep}{4pt}
\renewcommand{\arraystretch}{1.05}
\small
\begin{tabular}{l cc cc cc cc}
\toprule
& \multicolumn{2}{c}{\textbf{MinEval}} & \multicolumn{2}{c}{\textbf{MagicBrush}} & \multicolumn{2}{c}{\textbf{AnyBench}} & \multicolumn{2}{c}{\textbf{Emu-Edit}} \\
\cmidrule(lr){2-3}\cmidrule(lr){4-5}\cmidrule(lr){6-7}\cmidrule(lr){8-9}
Method & HPS & IR & HPS & IR & HPS & IR & HPS & IR \\
\midrule
FLUX.1~Kontext-dev (Base) & 0.212          & $-$0.80 & 0.224 & $-$0.51 & 0.215 & $-$0.58 & 0.208 & $-$0.78 \\
InstructPix2Pix           & \textbf{0.220} & $-$0.67 & 0.213 & $-$0.96 & 0.212 & $-$0.85 & 0.199 & $-$1.08 \\
MagicBrush model          & 0.211          & $-$0.84 & 0.211 & $-$0.86 & 0.205 & $-$0.88 & 0.199 & $-$1.00 \\
OmniGen2                  & 0.211          & $-$0.83 & 0.221 & $-$0.59 & 0.215 & $-$0.68 & 0.207 & $-$0.77 \\
UniWorld-v1              & 0.203          & $-$0.92 & 0.206 & $-$0.90 & 0.204 & $-$0.87 & 0.196 & $-$0.99 \\
ICEdit                    & 0.209          & $-$0.82 & 0.212 & $-$0.70 & 0.208 & $-$0.81 & 0.199 & $-$0.89 \\
BAGEL                     & 0.208          & $-$0.84 & 0.214 & $-$0.59 & 0.208 & $-$0.75 & 0.207 & $-$0.72 \\
\midrule
\rowcolor[gray]{0.92}
\methodname{} (ours)      & \textbf{0.220} & \textbf{$-$0.63} & \textbf{0.233} & \textbf{$-$0.29} & \textbf{0.221} & \textbf{$-$0.49} & \textbf{0.214} & \textbf{$-$0.62} \\
\bottomrule
\end{tabular}
\end{table}

\subsection{Per-Edit-Type Breakdown on AnyBench}
\label{app:per_edit_type}

Tables~\ref{tab:per_edit_type_hps_app} and~\ref{tab:per_edit_type_ir_app} report the per-edit-type HPSv2 and ImageReward breakdown on AnyBench (50 prompts per edit type), complementing the EditScore table in the main paper (Table~\ref{tab:per_edit_type}). The methods, edit-type columns, and best/second-best ($^*$) marking convention all match Table~\ref{tab:per_edit_type}.

\begin{table}[t]
\centering
\caption{Per-edit-type \textsc{HPSv2} on AnyBench (50 prompts per edit type), complementing Table~\ref{tab:per_edit_type} in the main paper. Edit-type columns and method order match Table~\ref{tab:per_edit_type}. Higher is better. The best score in each column is shown in \textbf{bold}, and the second-best score is marked with $^*$.}
\label{tab:per_edit_type_hps_app}
\setlength{\tabcolsep}{3pt}
\renewcommand{\arraystretch}{1.0}
\scriptsize
\resizebox{\linewidth}{!}{%
\begin{tabular}{l cccccccccccccccccc}
\toprule
\multirow{2}{*}{Method} & \multicolumn{18}{c}{Edit type} \\
\cmidrule(lr){2-19}
& act. & add & app. & bg. & color & count & impl. & mat. & move & outp. & rel. & rem. & repl. & resize & rot. & style & text & tone \\
\midrule
Base      & 0.266$^*$ & 0.266$^*$ & 0.245$^*$ & 0.210 & 0.223 & 0.177 & 0.240 & 0.221 & 0.210 & 0.191$^*$ & 0.215 & 0.169 & 0.234$^*$ & 0.221 & 0.183$^*$ & 0.191 & 0.233 & 0.177 \\
\rowcolor[gray]{0.92}
\methodname{} & \textbf{0.269} & \textbf{0.268} & \textbf{0.255} & 0.218 & 0.234 & 0.182 & \textbf{0.251} & 0.227 & 0.219$^*$ & \textbf{0.193} & \textbf{0.219} & 0.165 & \textbf{0.241} & 0.226$^*$ & \textbf{0.184} & \textbf{0.202} & 0.231 & 0.187 \\
IP2P      & 0.247 & 0.229 & 0.222 & 0.198 & 0.230 & \textbf{0.203} & 0.226 & 0.230$^*$ & \textbf{0.221} & 0.170 & 0.218$^*$ & \textbf{0.207} & 0.222 & \textbf{0.230} & 0.173 & 0.190 & 0.227 & 0.166 \\
MagicBr.  & 0.233 & 0.240 & 0.228 & 0.222$^*$ & 0.221 & 0.181 & 0.216 & 0.217 & 0.211 & 0.165 & 0.203 & 0.193$^*$ & 0.217 & 0.216 & 0.161 & 0.175 & 0.208 & 0.190$^*$ \\
OmniGen2  & 0.261 & 0.252 & 0.242 & \textbf{0.222} & 0.235$^*$ & 0.172 & 0.247$^*$ & \textbf{0.234} & 0.187 & 0.188 & 0.205 & 0.173 & 0.229 & 0.217 & 0.176 & 0.200 & \textbf{0.236} & \textbf{0.190} \\
UniWorld-v1 & 0.250 & 0.249 & 0.226 & 0.196 & 0.229 & 0.168 & 0.232 & 0.227 & 0.196 & 0.175 & 0.204 & 0.168 & 0.220 & 0.198 & 0.175 & 0.201$^*$ & 0.195 & 0.167 \\
ICEdit    & 0.233 & 0.241 & 0.233 & 0.183 & \textbf{0.237} & 0.188$^*$ & 0.213 & 0.227 & 0.203 & 0.180 & 0.210 & 0.192 & 0.225 & 0.203 & 0.172 & 0.194 & 0.229 & 0.184 \\
BAGEL     & 0.266 & 0.260 & 0.232 & 0.203 & 0.226 & 0.168 & 0.247 & 0.220 & 0.193 & 0.175 & 0.193 & 0.173 & 0.214 & 0.204 & 0.175 & 0.193 & 0.234$^*$ & 0.176 \\
\bottomrule
\end{tabular}}
\end{table}

\begin{table}[t]
\centering
\caption{Per-edit-type \textsc{ImageReward} on AnyBench (50 prompts per edit type), complementing Table~\ref{tab:per_edit_type} in the main paper. Edit-type columns and method order match Table~\ref{tab:per_edit_type}. Higher is better. The best score in each column is shown in \textbf{bold}, and the second-best score is marked with $^*$.}
\label{tab:per_edit_type_ir_app}
\setlength{\tabcolsep}{3pt}
\renewcommand{\arraystretch}{1.0}
\scriptsize
\resizebox{\linewidth}{!}{%
\begin{tabular}{l cccccccccccccccccc}
\toprule
\multirow{2}{*}{Method} & \multicolumn{18}{c}{Edit type} \\
\cmidrule(lr){2-19}
& act. & add & app. & bg. & color & count & impl. & mat. & move & outp. & rel. & rem. & repl. & resize & rot. & style & text & tone \\
\midrule
Base      & 0.42$^*$ & \textbf{0.79} & 0.27$^*$ & $-$0.81 & $-$0.15 & $-$1.33 & 0.05 & $-$0.79 & $-$0.46 & $-$1.12 & $-$0.49 & $-$1.76 & \textbf{$-$0.66} & \textbf{$-$0.44} & $-$1.57$^*$ & $-$1.15 & $-$0.09 & $-$1.18 \\
\rowcolor[gray]{0.92}
\methodname{} & \textbf{0.56} & 0.78$^*$ & \textbf{0.42} & $-$0.59 & 0.08$^*$ & $-$1.21 & \textbf{0.32} & $-$0.54$^*$ & \textbf{$-$0.35} & $-$0.99$^*$ & \textbf{$-$0.40} & $-$1.80 & $-$0.76$^*$ & $-$0.47$^*$ & \textbf{$-$1.55} & \textbf{$-$1.02} & $-$0.21 & \textbf{$-$1.07} \\
IP2P      & $-$0.43 & $-$0.61 & $-$0.86 & $-$1.16 & $-$0.50 & \textbf{$-$0.81} & $-$0.53 & $-$0.74 & $-$0.38$^*$ & $-$1.44 & $-$0.44$^*$ & \textbf{$-$1.10} & $-$1.18 & $-$0.55 & $-$1.73 & $-$1.19 & \textbf{0.02} & $-$1.61 \\
MagicBr.  & $-$0.36 & $-$0.14 & $-$0.49 & $-$0.58$^*$ & $-$0.60 & $-$1.22 & $-$0.55 & $-$1.01 & $-$0.67 & $-$1.43 & $-$0.63 & $-$1.34$^*$ & $-$1.12 & $-$0.81 & $-$1.88 & $-$1.27 & $-$0.62 & $-$1.15$^*$ \\
OmniGen2  & 0.05 & 0.34 & 0.02 & \textbf{$-$0.54} & $-$0.02 & $-$1.48 & $-$0.05 & \textbf{$-$0.51} & $-$1.05 & \textbf{$-$0.99} & $-$0.61 & $-$1.91 & $-$0.94 & $-$0.59 & $-$1.70 & $-$1.06$^*$ & $-$0.07$^*$ & $-$1.15$^*$ \\
UniWorld-v1 & 0.14 & 0.25 & $-$0.35 & $-$0.96 & $-$0.26 & $-$1.48 & $-$0.46 & $-$0.87 & $-$0.91 & $-$1.28 & $-$0.66 & $-$1.78 & $-$0.90 & $-$0.98 & $-$1.74 & $-$1.07 & $-$0.76 & $-$1.57 \\
ICEdit    & $-$0.68 & 0.06 & $-$0.26 & $-$1.31 & \textbf{0.10} & $-$1.09$^*$ & $-$0.63 & $-$0.60 & $-$0.81 & $-$1.24 & $-$0.55 & $-$1.47 & $-$0.91 & $-$0.84 & $-$1.68 & $-$1.23 & $-$0.23 & $-$1.18 \\
BAGEL     & 0.23 & 0.72 & $-$0.16 & $-$0.89 & $-$0.16 & $-$1.44 & 0.20$^*$ & $-$0.63 & $-$0.84 & $-$1.29 & $-$0.86 & $-$1.84 & $-$1.16 & $-$0.87 & $-$1.75 & $-$1.33 & $-$0.11 & $-$1.35 \\
\bottomrule
\end{tabular}}
\end{table}

\section{Training Settings}
\label{app:training_settings}

We list the key hyperparameters used to train \methodname{} in Tables~\ref{tab:hyperparams} and~\ref{tab:hyperparams_cont}. All FLUX.1 Kontext-dev runs share this configuration; ablations in Sec.~\ref{sec:exp_ablation} change only their specified reward components.

\begin{table}[H]
\centering
\caption{Key hyperparameters of \methodname{} fine-tuning on FLUX.1~Kontext-dev.}
\label{tab:hyperparams}
\small
\setlength{\tabcolsep}{8pt}
\renewcommand{\arraystretch}{1.05}
\begin{tabular}{lc}
\toprule
\textbf{Parameter} & \textbf{Setting} \\
\midrule
\multicolumn{2}{c}{\emph{Optimizer}} \\
\midrule
Optimizer                  & AdamW \\
Learning rate              & $1\!\times\!10^{-5}$ \\
Weight decay               & $1\!\times\!10^{-4}$ \\
Gradient-norm clip         & $1.0$ \\
PPO ratio clip             & $5\!\times\!10^{-3}$ \\
Precision                  & bf16 \\
\midrule
\multicolumn{2}{c}{\emph{Sampler (training)}} \\
\midrule
Resolution                 & $720\!\times\!720$ \\
Denoising steps $T$        & $25$ \\
Flow-match shift           & $3$ \\
Stochasticity $\eta$       & $0.7$ \\
Group size $N$ (candidates per prompt) & $16$ \\
\midrule
\multicolumn{2}{c}{\emph{MixGRPO}} \\
\midrule
Sliding window             & first $60\%$ of schedule \\
\texttt{iters\_per\_group} & $20$ \\
\texttt{group\_size} (sliding-window size, in steps) & $4$ \\
Progressive overlap        & enabled \\
\bottomrule
\end{tabular}
\end{table}

\begin{table}[H]
\centering
\caption{Reward, compute, and inference settings for FLUX.1~Kontext-dev experiments.}
\label{tab:hyperparams_cont}
\small
\setlength{\tabcolsep}{8pt}
\renewcommand{\arraystretch}{1.05}
\begin{tabular}{lc}
\toprule
\textbf{Parameter} & \textbf{Setting} \\
\midrule
\multicolumn{2}{c}{\emph{Reward}} \\
\midrule
Channel-1 weight $w_1$     & $0.3$ \\
Channel-2 weight $w_2$     & $0.7$ \\
Auditor VLM                & Qwen3-VL-30B-A3B-Instruct~\citep{qwen3} \\
Aggregator LLM             & Qwen3-30B-A3B-Instruct-2507~\citep{qwen3} \\
\midrule
\multicolumn{2}{c}{\emph{Compute}} \\
\midrule
Hardware                   & $8\!\times\!\text{A100-80GB}$ (single node) \\
Per-GPU prompts/step       & $1$ \\
Gradient accumulation      & $4$ \\
Effective batch            & $32$ prompts/step \\
Total optimizer steps      & $230$ \\
Wall-clock                 & $\approx$300 GPU-hr \\
\midrule
\multicolumn{2}{c}{\emph{Sampler (inference)}} \\
\midrule
Resolution                 & $720\!\times\!720$ \\
Denoising steps $T$        & $25$ \\
Guidance scale             & $3.5$ \\
Stochasticity $\eta$       & $0$ \\
Seed                       & $42$ (deterministic) \\
\bottomrule
\end{tabular}
\end{table}

\section{Training Data Construction}
\label{app:data_construction}

This appendix expands the three-stage data pipeline summarized in Sec.~\ref{sec:exp_setup}. Stage (i) synthesizes atomic instructions, stage (ii) filters them through a structured VLM audit, and stage (iii) executes them with the un-tuned base editor and rejects instructions whose candidate score distributions are degenerate.

\subsection{Atomic Edit Types}
\label{app:data_edit_types}

Every instruction is constrained to one of six atomic edit types, each with a fixed surface pattern and a closed list of allowed verbs:
\begin{samepage}
\begin{itemize}[leftmargin=1.2em,topsep=2pt plus 0pt minus 0pt,itemsep=1pt plus 0pt minus 0pt,parsep=0pt plus 0pt minus 0pt]
    \item \texttt{add} --- ``\emph{add/insert/put/place} \texttt{<object>} \emph{to/at} \texttt{<location>}''.
    \item \texttt{remove} --- ``\emph{remove/delete/erase} \texttt{<object>}''.
    \item \texttt{move} --- ``\emph{move/relocate} \texttt{<object>} \emph{from} \texttt{<src>} \emph{to} \texttt{<dst>}''.
    \item \texttt{replace} --- ``\emph{replace/substitute/swap} \texttt{<old object>} \emph{with} \texttt{<new object>}''.
    \item \texttt{appearance\_change} --- ``\emph{change/make/paint} \texttt{<object>}'s \texttt{<attribute>} \emph{to} \texttt{<value>}'' (color, texture, material).
    \item \texttt{action\_change} --- ``\emph{make/have} \texttt{<animate object>} \texttt{<action>}'' (action must be species-appropriate).
\end{itemize}
\end{samepage}
Instructions whose leading verb conflicts with the declared edit type (e.g., a \texttt{remove}-style verb under \texttt{add}) are rejected at parse time before any VLM call.

\subsection{Instruction Generation and Validation}
\label{app:data_prompts}

For each source image $\bm{x}_{\text{src}}$ and edit type $t$, the VLM is invoked in one of two modes. When a draft instruction $c_0$ is available (e.g., from a prior pass), the \emph{validation} prompt $\mathsf{P}_{\text{val}}$ asks the VLM to audit $c_0$ against a fixed checklist and, when fixable, return a corrected instruction. When no draft is available, or after $5$ failed validation rounds, the \emph{generation} prompt $\mathsf{P}_{\text{gen}}$ asks the VLM to propose a fresh instruction from scratch and self-audit it under the same checklist. To improve diversity, the generation call is launched in parallel with $5$ different temperatures $\{0.6, 0.7, 0.8, 0.9, 1.0\}$ and the first candidate that passes the checklist is kept. Both prompts share the same JSON output schema; the validation variant adds an explicit \texttt{instruction} field for the corrected wording, the generation variant a \texttt{generated\_instruction} field. We show the validation variant below; the generation variant is identical except that the \texttt{<instruction\_to\_validate>} block is removed and Step 0 is replaced by ``select a single, high-confidence target object''.

\begin{tcolorbox}[breakable, title={$\mathsf{P}_{\text{val}}(\bm{x}_{\text{src}},\,c_0,\,t)$ (instruction-filtering prompt)}]
\small
\textbf{Role:} You are validating an image editing instruction. Decide whether the instruction is accurate, complete, and feasible for the given source image.

\medskip
\textbf{Inputs:}
\begin{itemize}
    \item \textbf{Source image:} \texttt{\{$\bm{x}_{\text{src}}$\}}
    \item \textbf{Instruction to validate:} \texttt{\{$c_0$\}}
    \item \textbf{Edit type:} \texttt{\{$t$\}} (one of \texttt{add}, \texttt{remove}, \texttt{move}, \texttt{replace}, \texttt{appearance\_change}, \texttt{action\_change}).
    \item \textbf{Required pattern} for $t$: as defined in App.~\ref{app:data_edit_types}.
\end{itemize}

\textbf{Step 1: Single-object check.} Does the instruction target exactly one object? Reject phrasings such as ``and'', ``both'', ``all'', or plural targets.

\medskip
\textbf{Step 2: Object verification.} Look at the source image and decide whether the referenced object actually exists and is correctly identified (i.e., not confused with a similar-looking object).

\medskip
\textbf{Step 3: Completeness check.} Verify that all slots required by the edit type are filled (e.g., \texttt{replace} requires both old and new object; \texttt{move} requires both source and target locations; \texttt{add} requires both object and location).

\medskip
\textbf{Step 4: State-change check.} Compare the current state of the target with the state requested by the instruction. If they already match, the edit is trivial and must be rejected.

\medskip
\textbf{Step 5: Feasibility check.} Decide whether the requested change is physically plausible; for \texttt{action\_change}, additionally check that the action is appropriate for the species of the target.

\medskip
\textbf{Output format.} Return a single JSON object and nothing else:
\begin{verbatim}
{
  "parsed_instruction": {
    "action": "<verb>", "target": "<single object>", "details": "<...>"
  },
  "single_object_check":   {"targets_single_object": true|false,
                            "violation_found": "<...>"},
  "object_verification":   {"object_exists": true|false,
                            "object_correctly_identified": true|false,
                            "what_i_actually_see": "<...>"},
  "completeness_check":    {"all_required_fields_present": true|false,
                            "missing_fields": ["<...>"]},
  "state_verification":    {"current_state": "<...>",
                            "requested_state": "<...>",
                            "states_are_different": true|false},
  "feasibility_check":     {"physically_possible": true|false,
                            "species_appropriate": true|false|null,
                            "issue": "<... or 'none'>"},
  "overall_result": {
    "feasible": true|false,
    "instruction": "<original if valid, corrected if fixable, else empty>",
    "failed_checks": ["<names of any failed checks>"],
    "reason": "<brief summary>"
  }
}
\end{verbatim}
\end{tcolorbox}

An instruction is accepted only if every check returns \texttt{true} and the \texttt{instruction} field, after pipeline-level verb-vs-edit-type re-check, is non-empty. Otherwise the failed-check names are fed back to the VLM as a recheck context for up to $5$ rounds, after which the pipeline falls back to parallel generation as described above. Instructions that never pass either route are discarded.

\subsection{Per-edit-type guidance injected into \texorpdfstring{$\mathsf{P}_{\text{val}}$}{P\_val} and \texorpdfstring{$\mathsf{P}_{\text{gen}}$}{P\_gen}}
\label{app:data_per_type_prompts}

Both $\mathsf{P}_{\text{val}}$ and $\mathsf{P}_{\text{gen}}$ inject two type-specific blocks under the placeholder \emph{``Required pattern for $t$''}: a \textbf{type-specific guidance} block (what makes a valid instruction of this type) and an \textbf{evaluation-alignment} block (which verifiable criteria the downstream agentic reward model will check). The six blocks below are summaries of the strings we actually inject for each $t$.

\begin{tcolorbox}[title={Edit type \texttt{add}}, fonttitle=\small\bfseries, fontupper=\small]
\textbf{Pattern.}\ \texttt{Add [specific object with attributes] to/in [precise location]}. \textbf{Allowed verbs:} \emph{add, put, place, insert}.

\medskip
\textbf{Type-specific guidance.} Introduce a \emph{new} object that does not yet exist in the image. Specify (i) the object: type, color, size (e.g., ``a large red balloon''); (ii) a precise location with confirmed empty space. Reject if the candidate location lacks empty space sufficient for the object's size, since adding the object would otherwise displace existing content and create side effects.

\medskip
\textbf{Evaluation alignment.} The downstream reward model checks: \emph{presence} (the new object appears), \emph{quantity} (count matches; always state quantity explicitly), \emph{position} (at the specified location), \emph{attributes} (color/size/type match), and \emph{addition $\neq$ replacement} (existing objects must remain). Use \emph{add/place}, never \emph{replace/swap}. Prefer empty-space targets so the evaluator can confirm no displacement.
\end{tcolorbox}

\begin{tcolorbox}[title={Edit type \texttt{remove}}, fonttitle=\small\bfseries, fontupper=\small]
\textbf{Pattern.}\ \texttt{Remove the [color/feature] [object] [position disambiguator]}. \textbf{Allowed verbs:} \emph{remove, delete, erase}.

\medskip
\textbf{Type-specific guidance.} Remove an object that clearly exists and can be uniquely identified. Confirm the object is visible, add a disambiguator when multiple similar objects exist, and verify the identification is correct (not a similar-looking object, e.g., sheep mistaken for bull).

\medskip
\textbf{Evaluation alignment.} The downstream reward model checks: \emph{true removal} (the object must be completely invisible; blurring, painting-over, or covering counts as failure) and \emph{disambiguation} (the evaluator must be able to locate the exact target). Use \emph{remove/delete/erase}, never \emph{hide/cover/obscure} (which imply the object is still present). Add enough disambiguators (color, position, size) when multiple similar objects exist.
\end{tcolorbox}

\begin{tcolorbox}[title={Edit type \texttt{move}}, fonttitle=\small\bfseries, fontupper=\small]
\textbf{Pattern.}\ \texttt{Move the [object] from [current position] to [different target position]}. \textbf{Allowed verbs:} \emph{move, shift, relocate}.

\medskip
\textbf{Type-specific guidance.} Relocate an existing object to a clearly different position. Verify (i) current position (where the object \emph{is now}) and (ii) target position (where it should \emph{go}, must be different from current). Use canonical position equivalences (\texttt{LEFT}=``left side'', \texttt{CENTER}=``middle'', etc.) when checking that source $\neq$ destination.

\medskip
\textbf{Evaluation alignment.} The downstream reward model checks: \emph{destination} (object present at the new position), \emph{source clear} (gone from the original; no residual or ghost), \emph{no duplication} (object must not appear at both positions), \emph{identity} (same object preserved). Always specify both ``from [source]'' and ``to [destination]'' so the evaluator can verify both positions; vague phrasing such as ``move the cup over'' fails because the evaluator cannot confirm source-clear.
\end{tcolorbox}

\begin{tcolorbox}[title={Edit type \texttt{replace}}, fonttitle=\small\bfseries, fontupper=\small]
\textbf{Pattern.}\ \texttt{Replace the [existing object] with [specific new object]}. \textbf{Allowed verbs:} \emph{replace, substitute, swap}.

\medskip
\textbf{Type-specific guidance.} Both the old and the new object must be specified explicitly. The old object must be visible in the source image; the new object should be similar in size/shape so it can fit naturally at the same location without major adjustments to the surrounding scene.

\medskip
\textbf{Evaluation alignment.} The downstream reward model checks: \emph{A gone} (old object completely absent), \emph{B present} (new object appears at A's original location), and \emph{no coexistence} (A and B must not both be visible; coexistence = failure). Name the old object precisely so the evaluator can confirm absence. Use ``replace A with B'', never just ``add B'' (which implies A stays).
\end{tcolorbox}

\begin{tcolorbox}[title={Edit type \texttt{appearance\_change}}, fonttitle=\small\bfseries, fontupper=\small]
\textbf{Pattern.}\ \texttt{Change the [object]'s [attribute] from [current value] to [new value]} (color, material, texture). \textbf{Allowed verbs:} \emph{change, modify, turn, make}.

\medskip
\textbf{Type-specific guidance.} Modify a single appearance attribute of an existing object. Verify (i) the object exists, (ii) its current attribute value is identified accurately, and (iii) the target value is different from the current one. Reject when current value already equals target (e.g., ``change the stop sign to red''), or when the named attribute does not exist on the object (e.g., ``change the saddle'' on an unsaddled horse).

\medskip
\textbf{Evaluation alignment.} The downstream reward model checks: \emph{identity preserved} (object keeps its shape, position, and category), \emph{attribute matches} (the specified attribute reaches the target value), and \emph{only the specified attribute changes}. Be specific about which attribute is being changed (``color from red to blue'', not ``make it different''); vague instructions cause false failures because the evaluator cannot determine the intended boundary.
\end{tcolorbox}

\begin{tcolorbox}[title={Edit type \texttt{action\_change}}, fonttitle=\small\bfseries, fontupper=\small]
\textbf{Pattern.}\ \texttt{Make the [specific subject] [new action] instead of [current action]}. \textbf{Allowed verbs:} \emph{make, have, change}.

\medskip
\textbf{Type-specific guidance.} Change a subject's pose, action, or expression to something different and \emph{species-appropriate}. Use species-appropriate action vocabularies (humans: sit/stand/walk/wave/lie down; quadrupeds: lie down/walk/turn head/graze, but not ``stand up'' since they are already standing on four legs; birds: perch/fly/spread wings/hop). Reject when current action equals target (e.g., ``make the person sit down'' on someone already sitting).

\medskip
\textbf{Evaluation alignment.} The downstream reward model checks: \emph{correct subject} (the right person/animal performs the action; disambiguate when multiple subjects exist), \emph{action visible} (target action is clearly visible in a static frame), \emph{subject identity preserved}. Choose actions that are visually verifiable in a single frame; avoid temporally extended actions like ``start walking'' or ``begin to speak'' that require motion to verify.
\end{tcolorbox}

\subsection{Edit-Level Rejection Sampling}
\label{app:data_reject}

The instructions that survive App.~\ref{app:data_prompts} are then filtered at the edit level. For every $(\bm{x}_{\text{src}}, c, t)$ triplet we sample a group of $N{=}16$ candidate edits $\mathcal{G}{=}\{\bm{x}_i\}_{i=1}^{N}$ from the un-tuned base editor (FLUX.1~Kontext-dev) at the training resolution of $720{\times}720$, with $T{=}25$ denoising steps and i.i.d.\ noise seeds; this matches the GDPO group used during RL fine-tuning (Sec.~\ref{sec:exp_impl}). Each candidate is then scored by the same three-step agentic reward pipeline used at training time: audit, group-level rubric aggregation, and rubric-anchored verification (Eqs.~\eqref{eq:audit}--\eqref{eq:rewards}), yielding the same two reward channels we optimize against:
\begin{equation*}
r^{(1)}_i \;=\; -\,|\mathcal{U}_i|, \qquad r^{(2)}_i \;=\; -\,|\mathcal{O}_i|,
\end{equation*}
that is, the negative counts of verified unimplemented and unintended issues for candidate $\bm{x}_i$.

To align with the GDPO advantage estimator, which is identically zero on a group whose rewards are constant, we filter on the variation of these two channels within $\mathcal{G}$. Concretely, we run a two-pass filter:
\begin{itemize}[leftmargin=1.2em,topsep=2pt,itemsep=1pt]
    \item \emph{Pass 1 (unimplemented channel).} Drop the group if more than half of its candidates have $|\mathcal{U}_i|{=}0$, i.e., if $\#\{i: |\mathcal{U}_i|{=}0\} > N/2$. When a majority of candidates already realize the instruction $c$ in full, the channel-1 reward $r^{(1)}_i$ is dominated by a single mode at $0$, the within-group variance of the unimplemented signal collapses, and GDPO receives little advantage signal from this channel. Such groups are dropped regardless of their channel-2 distribution.
    \item \emph{Pass 2 (unintended channel).} Among groups that pass Pass~1, drop those for which $|\mathcal{O}_i|$ is also constant across $i$, or for which $\mathrm{std}(|\mathcal{O}_{1{:}N}|){<}\sigma_{\min}$, or for which the per-group mean of $|\mathcal{O}_i|$ exceeds a maximum tolerated count $\mu_{\max}$. The first two conditions ensure the locality channel actually carries a within-group signal; the last condition discards prompts on which the base editor changes so much of the source image that the auditor would have to reason over a long, heterogeneous list of differences, which keeps per-item judgments focused and low-variance.
\end{itemize}

For the runs reported in Sec.~\ref{sec:exp} we use $\sigma_{\min}{=}0$ and $\mu_{\max}{=}999$, so that within Pass~2 only the constant-across-$i$ test on $|\mathcal{O}_i|$ is active (the std and mean tests are inert at these settings). In this default configuration the filter discards two kinds of groups: those whose unimplemented channel is dominated by zeros (Pass~1, majority-success), and those whose unintended channel is constant across the group (Pass~2, locality-channel constancy). Both are degenerate cases for the group-relative advantage estimator: in the first, the efficacy channel carries little ranking signal because most candidates already realize $c$ in full; in the second, the locality channel literally has zero within-group variance and contributes a zero advantage. Tightening $\sigma_{\min}$ or $\mu_{\max}$ yields the more aggressive variants used in our preliminary tuning runs but did not change the qualitative ranking. The retained triplets form the $\sim$10K atomic-edit training set used for RL fine-tuning.

\section{Prompt Templates}
\label{app:prompts}

We instantiate the three model invocations in Sec.~\ref{sec:method} with the following prompt templates. Each template is a fixed string with placeholders (in \texttt{\{curly braces\}}) for inputs. All templates instruct the model to return a JSON object with a fixed schema, which we parse to obtain the required outputs in Eqs.~\eqref{eq:audit}, \eqref{eq:rubric}, and \eqref{eq:verify}.

The audit prompt $\mathsf{P}_{\text{audit}}$ and the verification prompt $\mathsf{P}_{\text{verify}}$ are designed to share the same reasoning structure: both ask the VLM to parse the instruction, compare the source and edited images, and identify changes belonging to the two failure modes. They differ only in the output mode. The audit is open-ended, asking the VLM to generate the lists $\widetilde{\mathcal{U}}_i$ and $\widetilde{\mathcal{O}}_i$ from scratch. The verification is closed-form, asking the VLM to select, from each fixed group rubric, the items present in $\bm{x}_i$. This shared structure ensures the two stages apply the same notion of "unimplemented" and "unintended" change, and the rubric refinement loop only varies what the VLM is asked to \emph{produce}, not how it is asked to \emph{look}.

\subsection{Audit prompt \texorpdfstring{$\mathsf{P}_{\text{audit}}$}{P\_audit}}
\label{app:prompt_audit}

The audit prompt is given to the VLM $\Phi_{\text{V}}$ together with the source image $\bm{x}_{\text{src}}$, the candidate edit $\bm{x}_i$, and the instruction $c$, and produces the raw lists $(\widetilde{\mathcal{U}}_i, \widetilde{\mathcal{O}}_i)$.

\begin{tcolorbox}[breakable, title={$\mathsf{P}_{\text{audit}}(\bm{x}_{\text{src}},\,c,\,\bm{x}_i)$}]
\small
\textbf{Role:} You are a careful visual auditor. You will be shown two images and a natural-language editing instruction. Your task is to identify two kinds of discrepancies between the instruction and what the edited image actually shows.

\medskip
\textbf{Inputs:}
\begin{itemize}
    \item \textbf{Source image:} \texttt{\{$\bm{x}_{\text{src}}$\}}
    \item \textbf{Edited image:} \texttt{\{$\bm{x}_i$\}}
    \item \textbf{Editing instruction:} \texttt{\{$c$\}}
\end{itemize}

\textbf{Step 1: Parse the instruction.}
Identify what the instruction explicitly asks to modify, what the resulting state should be, and what should remain unchanged.

\medskip
\textbf{Step 2: Compare the two images.}
Examine the source and edited image side by side and note all visible differences. For each difference, decide whether it falls inside or outside the scope of the instruction.

\medskip
\textbf{Step 3: Generate two lists.}
\begin{enumerate}
    \item \textbf{Unimplemented changes ($\widetilde{\mathcal{U}}_i$):} aspects of the instruction that the edited image fails to realize, including complete omission, partial execution, or incorrect application. Return an empty list if the instruction is fully realized.
    \item \textbf{Unintended changes ($\widetilde{\mathcal{O}}_i$):} visible modifications that fall outside the scope of the instruction. Include only changes you can actually see; do not speculate. Return an empty list if no such changes are present.
\end{enumerate}
Each entry must be a short, self-contained, visually verifiable description of a single change. Do not assign numerical scores. Do not summarize.

\medskip
\textbf{Output format.} Return a single JSON object and nothing else:
\begin{verbatim}
{
  "reasoning": "<your step-1 and step-2 analysis>",
  "unimplemented_changes": [
    "<short description>", ...
  ],
  "unintended_changes": [
    "<short description>", ...
  ]
}
\end{verbatim}
\end{tcolorbox}

\subsection{Aggregation prompt \texorpdfstring{$\mathsf{P}_{\text{agg}}$}{P\_agg}}
\label{app:prompt_agg}

The aggregation prompt is invoked once per candidate group. It receives the two audit collections in separate fields, applies the instruction filter only to unintended changes, and returns both canonical rubrics in one response.

\begin{tcolorbox}[breakable, title={$\mathsf{P}_{\text{agg}}(c,\,\widetilde{\mathcal{U}}_{1{:}N},\,\widetilde{\mathcal{O}}_{1{:}N})$}]
\small
\textbf{Role:} You are a rubric editor. Given an editing instruction and two collections of free-form issues identified across $N$ candidate edits, construct separate, deduplicated checklists for unimplemented and unintended changes. Keep the two failure modes distinct.

\medskip
\textbf{Inputs:}
\begin{itemize}
    \item \textbf{Editing instruction:} \texttt{\{$c$\}}
    \item \textbf{Raw unimplemented issues:} candidate-indexed lists $\{\widetilde{\mathcal{U}}_i\}_{i=1}^{N}$.
    \item \textbf{Raw unintended issues:} candidate-indexed lists $\{\widetilde{\mathcal{O}}_i\}_{i=1}^{N}$.
\end{itemize}

\textbf{Step 1: Filter against the instruction.} From the unintended collection only, remove descriptions of changes explicitly requested or directly implied by $c$. Do not apply this filter to unimplemented issues, which describe missing aspects of $c$.

\medskip
\textbf{Step 2: Consolidate each collection separately.} Merge descriptions of the same underlying issue, split descriptions that conflate distinct issues, normalize each entry to one atomic visually checkable failure, and drop vague, speculative, or non-visual entries. For example, ``the bow was not removed'' and ``the bow is still visible'' form one unimplemented issue; ``the wheels became silver'' and ``the rims changed color'' form one unintended issue when they describe the same change.

\medskip
\textbf{Step 3: Produce two rubrics.} Write each issue in neutral form so that it can be checked against any candidate. Assign IDs within each rubric. Do not merge an unimplemented issue with an unintended issue.

\medskip
\textbf{Output format.} Return a single JSON object and nothing else:
\begin{verbatim}
{
  "reasoning": "<filtering and consolidation rationale>",
  "unimplemented_rubric": [
    {"id": 1, "issue": "<canonical unimplemented issue>"}, ...
  ],
  "unintended_rubric": [
    {"id": 1, "issue": "<canonical unintended issue>"}, ...
  ]
}
\end{verbatim}
\end{tcolorbox}

The two returned arrays define $\mathcal{R}^{\mathcal{U}}_\mathcal{G}$ and $\mathcal{R}^{\mathcal{O}}_\mathcal{G}$, respectively.

\subsection{Verification prompt \texorpdfstring{$\mathsf{P}_{\text{verify}}$}{P\_verify}}
\label{app:prompt_verify}

The verification prompt is invoked once per candidate. It receives both group rubrics and returns separate Boolean verdicts for their issues in one response. As in the audit prompt, the VLM first parses the instruction and compares the source and edited images; verification then selects from the fixed rubrics rather than generating new issues.

\begin{tcolorbox}[breakable, title={$\mathsf{P}_{\text{verify}}(\bm{x}_{\text{src}},\,c,\,\bm{x}_i,\,\mathcal{R}^{\mathcal{U}}_\mathcal{G},\,\mathcal{R}^{\mathcal{O}}_\mathcal{G})$}]
\small
\textbf{Role:} You are a strict visual verifier. Given a source image, an editing instruction, an edited image, and two checklists of canonical issues, decide which issues are visibly present. Keep unimplemented and unintended changes separate.

\medskip
\textbf{Inputs:}
\begin{itemize}
    \item \textbf{Source image:} \texttt{\{$\bm{x}_{\text{src}}$\}}
    \item \textbf{Edited image:} \texttt{\{$\bm{x}_i$\}}
    \item \textbf{Editing instruction:} \texttt{\{$c$\}}
    \item \textbf{Unimplemented checklist:} $\mathcal{R}^{\mathcal{U}}_\mathcal{G}$, with issue IDs and descriptions.
    \item \textbf{Unintended checklist:} $\mathcal{R}^{\mathcal{O}}_\mathcal{G}$, with issue IDs and descriptions.
\end{itemize}

\textbf{Step 1: Parse the instruction.} Identify what $c$ asks to modify, the required result, and what should remain unchanged.

\medskip
\textbf{Step 2: Compare the images.} Examine the source and edited image side by side. Determine which requested aspects were realized and which visible differences fall outside the scope of $c$.

\medskip
\textbf{Step 3: Select from both checklists.} For each unimplemented item, mark \texttt{true} only if the named aspect of $c$ remains absent, partial, or incorrect. For each unintended item, mark \texttt{true} only if that off-target change is visible. When in doubt, mark \texttt{false}. Do not add issues or alter their wording.

\medskip
\textbf{Output format.} Return a single JSON object and nothing else:
\begin{verbatim}
{
  "reasoning": "<instruction and image comparison>",
  "unimplemented_verdicts": [
    {"id": 1, "present": true}, ...
  ],
  "unintended_verdicts": [
    {"id": 1, "present": false}, ...
  ]
}
\end{verbatim}
\end{tcolorbox}

The issues marked \texttt{true} in the two verdict arrays form $\mathcal{U}_i\subseteq\mathcal{R}^{\mathcal{U}}_\mathcal{G}$ and $\mathcal{O}_i\subseteq\mathcal{R}^{\mathcal{O}}_\mathcal{G}$. Consequently, one candidate group needs $N$ audit calls, one aggregation call, and $N$ verification calls, for $2N+1$ reward-generation calls.

\end{document}